%% file: main.tex
\documentclass{article}

     \usepackage[preprint]{neurips_2024}

\usepackage[utf8]{inputenc} %
\usepackage[T1]{fontenc}    %
\usepackage{url}            %
\usepackage{booktabs}       %
\usepackage{amsfonts}       %
\usepackage{nicefrac}       %
\usepackage{microtype}      %
\usepackage{xcolor}         %

\input{macros}
\usepackage{caption}
\usepackage{subcaption}
\usepackage{multirow}

\title{Bandits with Preference Feedback:\\
A Stackelberg Game Perspective}

\author{
Barna P{\'a}sztor$^{\star, 1,2}$ 
  \quad Parnian Kassraie$^{\star, 1}$
  \quad Andreas Krause$^{1, 2}$\\
  $^1$ETH Zurich \quad $^2$ETH AI Center\\
  \texttt{\{bpasztor, pkassraie, krausea\}@ethz.ch}
  }

\begin{document}

\maketitle
\def\thefootnote{$\star$}\footnotetext{Equal contribution. }\def\thefootnote{\arabic{footnote}}

\addtocontents{toc}{\protect\setcounter{tocdepth}{0}}
\input{sections/0_abstract}

\input{sections/1_introduction}

\input{sections/2_related_works}

\input{sections/3_problem_setting}

\input{sections/4_logistic_warmup}

\input{sections/5_main_result}

\input{sections/6_experiments}

\input{sections/7_conclusion}

 \begin{ack}
 We thank Scott Sussex for his thorough feedback.
This research was supported by the European Research Council (ERC) under the European Union’s Horizon 2020 research and Innovation Program Grant agreement No. 815943. Barna P{\'a}sztor was supported by an ETH AI Center doctoral fellowship, and Parnian Kassraie by a Google PhD Fellowship.\looseness-1
 \end{ack}

\bibliographystyle{plainnat}
\bibliography{refs}

\newpage
 \appendix
 \renewcommand*\contentsname{Contents of Appendix}
 \addtocontents{toc}{\protect\setcounter{tocdepth}{2}}
\tableofcontents
\input{appendices/1_logistic_proofs}
\input{appendices/2_dueling_proofs}

\input{appendices/3_ids_proof}
\input{appendices/4_additional_experiments}

\end{document}

%% file: macros.tex
\usepackage{natbib}
\usepackage{amsfonts,bm}
\usepackage{amssymb,mathrsfs, amsthm}
\usepackage{xspace}
\usepackage{enumitem}
\usepackage{lipsum}
\usepackage{xpatch}
\usepackage{mathtools}
\usepackage{dsfont}
\usepackage{pgf,tikz}
\usetikzlibrary{positioning,matrix}
\usepackage{caption}
\usepackage{graphicx}
\usepackage{xcolor}
\usepackage{parskip}
\definecolor{parn}{rgb}{0.27, 0.42, 0.81}
\usepackage[hidelinks]{hyperref}
\hypersetup{
    colorlinks=true,
    linkcolor=parn,
    urlcolor=parn,
    citecolor=parn,
    anchorcolor=parn}

 \usepackage[capitalize,noabbrev]{cleveref}

\theoremstyle{plain}
\newtheorem{theorem}{Theorem}
\newtheorem{proposition}[theorem]{Proposition}
\newtheorem{lemma}[theorem]{Lemma}
\newtheorem{corollary}[theorem]{Corollary}
\theoremstyle{definition}

\theoremstyle{remark}

\usepackage{algorithm}
\usepackage{algpseudocode}

\def\vtheta{{\bm{\theta}}}

\def\valpha{{\bm{\alpha}}}

\def\vphi{{\bm{\phi}}}

\def\vpsi{{\bm{\psi}}}

\def\vg{{\bm{g}}}

\def\vk{{\bm{k}}}

\def\vx{{\bm{x}}}

\def\vz{{\bm{z}}}

\def\mI{{\bm{I}}}

\def\calE{{\mathcal{E}}}
\def\calF{{\mathcal{F}}}

\def\calH{{\mathcal{H}}}

\def\calL{{\mathcal{L}}}
\def\calM{{\mathcal{M}}}
\def\calN{{\mathcal{N}}}
\def\calO{{\mathcal{O}}}
\def\calP{{\mathcal{P}}}

\def\calX{{\mathcal{X}}}

\def\calZ{{\mathcal{Z}}}

\def\sN{{\mathbb{N}}}

\def\sP{{\mathbb{P}}}

\def\sR{{\mathbb{R}}}

\newcommand{\Exp}{\mathbb{E}}

\DeclareMathOperator*{\argmax}{arg\,max}
\DeclareMathOperator*{\argmin}{arg\,min}
\newcommand{\sigmoid}{s}

\newcommand{\eqcolon}{\mathrel{\resizebox{\widthof{$\mathord{=}$}}{\height}{ $\!\!=\!\!\resizebox{1.2\width}{0.8\height}{\raisebox{0.23ex}{$\mathop{:}$}}\!\!$ }}}

\DeclarePairedDelimiter\abs{\lvert}{\rvert}%
\DeclarePairedDelimiter\norm{\lVert}{\rVert}%

\makeatletter
\let\oldabs\abs
\def\abs{\@ifstar{\oldabs}{\oldabs*}}
\let\oldnorm\norm
\def\norm{\@ifstar{\oldnorm}{\oldnorm*}}
\makeatother

\def\algo{{\textsc{LGP-UCB}}\xspace}
\def\dalgo{{\textsc{MaxMinLCB}}\xspace}
\def\UCB{{\textsc{Ind-UCB}}\xspace}
\def\GPUCB{{\textsc{GP-UCB}}\xspace}
\def\LOGUCB{{\textsc{Log-UCB1}}\xspace}

\def\RUCB{{\textsc{RUCB}}\xspace}

\def\MaxInfo{{\textsc{MaxInP}}\xspace}
\def\MultiSBM{{\textsc{MultiSBM}}\xspace}
\def\IDS{{\textsc{IDS}}\xspace}
\def\Doubler{{\textsc{Doubler}}\xspace}

\def\POPBO{{\textsc{POP-BO}}\xspace}

\def\tf{{f}} %
\def\fhat{{f_t}}%
\def\kduel{k^{\mathrm{D}}}

\definecolor{hanblue}{rgb}{0.27, 0.42, 0.81}
\definecolor{deepred}{HTML}{900C3F}
\definecolor{deepgreen}{HTML}{2F6960}

\usepackage{pifont}
\usepackage{makecell}
\usepackage{multirow}

%% file: sections/0_abstract.tex
\begin{abstract}

Bandits with preference feedback present a powerful tool for optimizing unknown target functions when only pairwise comparisons are allowed instead of direct value queries. This model allows for incorporating human feedback into online inference and optimization and has been employed in systems for fine-tuning large language models.
The problem is well understood in simplified settings with linear target functions or over finite small domains that limit practical interest.
Taking the next step, we consider infinite domains and nonlinear (kernelized) rewards. In this setting, selecting a pair of actions is quite challenging and requires balancing exploration and exploitation at two levels: within the pair, and along the iterations of the algorithm.
We propose \dalgo, which emulates this trade-off as a zero-sum Stackelberg game, and chooses action pairs that are informative and yield favorable rewards. \dalgo consistently outperforms existing algorithms and satisfies an anytime-valid rate-optimal regret guarantee. This is due to our novel preference-based confidence sequences for kernelized logistic estimators.  \looseness-1%

\end{abstract}

%% file: sections/1_introduction.tex
\section{Introduction}
\looseness=-1
In standard bandit optimization, a learner repeatedly interacts with an unknown environment that gives numerical feedback on the chosen actions according to a utility function $f$.
However, in applications such as fine-tuning large language models, drug testing, or search engine optimization, the quantitative value of design choices or test outcomes are either not directly observable, or are known to be inaccurate, or systematically biased, e.g., if they are provided by human feedback \citep{casper2023open}.
A solution is to optimize for the target based on comparative feedback provided for a pair of queries, which is proven to be more robust to certain biases and uncertainties in the queries \citep{ji2023provable}.

\looseness=-1
Bandits with preference feedback, or {\em dueling} bandits, address this problem and propose strategies for choosing query/action pairs that yield a high utility over the horizon of interactions.
At the core of such algorithms is uncertainty quantification and inference for $f$ in regions of interest, which is closely tied to exploration and exploitation dilemma over a course of queries.
Observing only comparative feedback poses an additional challenge, as we now need to balance this trade-off {\em jointly} over two actions.
This challenge is further exacerbated when optimizing over vast or infinite action domains.
As a remedy, prior work often {\em grounds} one of the actions by choosing it either randomly or greedily, and tries to balance exploration-exploitation for the second action as a reaction to the first \citep{ailon2014reducing, zoghi2014relative, kirschner2021bias, mehta2023kernelized}. 
This approach works well for simple utility functions over low-dimensional domains, however does not scale to more complex problems.

\looseness=-1 
Aiming to solve this problem, we focus on continuous domains in the Euclidean vector space and complex utility functions that belong to the Reproducing Kernel Hilbert Space (RKHS) of a potentially non-smooth kernel.
We propose \dalgo, a sample-efficient algorithm that at every step chooses the actions {\em jointly}, by playing a zero-sum Stackelberg (a.k.a~Leader-Follower) game.
We choose the Lower Confidence Bound (LCB) of $f$ as the objective of this game which the Leader aims to maximize and the Follower to minimize.
The equilibrium of this game yields an action pair in which the first action is a favorable candidate to maximize $f$ and the second action is the strongest competitor against the first.
Our choice of using the LCB as the objective leads to robustness against uncertainty when selecting the first action. Moreover, it makes the second action an optimistic choice as a competitor, from its own perspective.
We observe empirically that this approach creates a natural exploration scheme, and in turn, yields a more sample-efficient algorithm compared to standard baselines. 

\looseness=-1 
Our game-theoretic strategy leads to an efficient bandit solver, if the LCB is a valid and tight lower bound on the utility function.
To this end, we construct a confidence sequence for $f$ given pairwise preference feedback, by modeling the noisy comparative observations with a logistic-type likelihood function.
Our confidence sequence is anytime valid and holds uniformly over the domain, 
under the assumption that $f$ resides in an RKHS.
We improve prior work by removing or relaxing assumptions on the utility while maintaining the same rate of convergence.
This result on preference-based confidence sequences may be of independent interest, as it targets the loss function that is typically used for Reinforcement Learning with Human Feedback.

\textbf{Contributions} Our main contributions are:
\begin{itemize}
    \item We propose a novel game-theoretic acquisition function for pairwise action selection with preference feedback.
    \item We construct preference-based confidence sequences for kernelized utility functions that are tight and anytime valid.
    \item Together this creates \dalgo, an algorithm for bandit optimization with preference feedback over continuous domains. \dalgo satisfies $\calO(\gamma_T\sqrt{T})$ regret, where $T$ is the horizon and $\gamma_T$ is the {\em information gain} of the kernel.
    \item We benchmark \dalgo over a set of standard optimization problems and consistently outperform the  common baselines from the literature.
\end{itemize}

%% file: sections/2_related_works.tex
\section{Related Work}
Learning with indirect feedback was first studied in the supervised preference learning setting \citep{ aiolli2004learning, chu2005preference}. Subsequently, online and sequential settings were considered, motivated by applications in which the feedback is provided in an online manner, e.g., by a human \citep{yue2012k, yue2009interactively, houlsby2011bayesian}. \citet{bengs2021preference} surveys this field comprehensively; here we include a brief background.

Referred to as dueling bandits, a rich body of work considers (finite) multi-armed domains and learns a preference matrix specifying the relation among the arms.
Such work often relies on efficient sorting or tournament systems based on the frequency of wins for each action \citep{jamieson2011active, zoghi2014relative2, komiyama2015optimal,wu2016double,falahatgar2017maximum}. Rather than jointly selecting the arms, such strategies often simplify the problem by selecting one at random \citep{zoghi2014relative, zimmert2018factored}, greedily \citep{chen2017dueling}, or from the set of previously selected arms \citep{ailon2014reducing}. 
In contrast, we jointly optimize both actions by choosing them as the equilibrium of a two-player zero-sum Stackelberg game, enabling a more efficient exploration/exploitation trade-off. 

\looseness-1
The multi-armed dueling setting, which is reducible to multi-armed bandits \citep{ailon2014reducing}, naturally fails to scale to infinite compact domains, since regularity among ``similar'' arms is not exploited.
To go beyond finite domains, {\em utility-based} dueling bandits consider an unknown latent function that captures the underlying preference, instead of relying on a preference matrix. The preference feedback is then modeled as the difference in the utility of two chosen actions passed through a link function. 
Early work is limited to convex domains and imposes strong regularity assumptions
\citep{yue2009interactively, kumagai2017regret}. 
These assumptions are then relaxed to general compact domains if the utility function is linear \citep{dudik2015contextual,saha2021optimal,saha2022efficient}.
Constructing valid confidence sets from comparative feedback is a challenging task. However, it is strongly related to uncertainty quantification with direct logistic feedback, which is extensively analyzed by the literature on logistic and generalized linear bandits 
\citep{filippi2010parametric, faury2020improved,foster2018contextual,beygelzimer2019bandit,faury2022jointly,lee2024improved}.

\looseness=-1
Preference-based bandit optimization with linear utility functions is well understood and is even extended to reinforcement learning with preference feedback on trajectories \citep{pacchiano2021dueling,zhan2023provable,zhu2024principled,ji2023provable,munos2023nash}.
However, such approaches have limited practical interest, since they cannot capture real-world problems with complex nonlinear utility functions.
Alternatively, Reproducing Kernel Hilbert Spaces (RKHS) provide a rich model class for the utility, e.g., if the chosen kernel is universal. 
Many have proposed heuristic algorithms for bandits and Bayesian optimization in kernelized settings, albeit without providing theoretical guarantees \cite{brochu2010tutorial,gonzalez2017preferential,sui2017multi,tucker2020preference,mikkola2020projective,takeno2023towards}. 

There have been attempts to prove convergence of kernelized algorithms for preference-based bandits \citep{xu2020zeroth, kirschner2021bias, mehta2023kernelized, mehta2023sample}. Such works employ a regression likelihood model which requires them to assume that both the utility and the probability of preference, as a function of actions, lie in an RKHS. In doing so, they use a regression model for solving a problem that is inherently a classification. While the model is valid, it does not result in a sample-efficient algorithm.
In contrast, we use a kernelized logistic negative log-likelihood loss to infer the utility function, and provide confidence sets for its minimizer. 
In a concurrent work, \citet{xu2024principled} also consider the kernelized logistic likelihood model and propose a variant of the \MultiSBM algorithm \citep{ailon2014reducing} which uses likelihood ratio confidence sets. The theoretical approach and resulting algorithm bear significant differences, and the regret guarantee has a strictly worse dependency on the time horizon $T$, by a factor of $T^{1/4}$. This is discussed in more detail in \cref{sec:kernelized_bandits_preference_feedback}.\looseness-1

%% file: sections/3_problem_setting.tex
\section{Problem Setting}\label{sec:problem}
Consider an agent which repeatedly interacts with an environment: at step $t$ the agent selects two actions $\vx_t,\, \vx'_t \in \calX$ and only observes stochastic binary feedback $y_t \in \{0,1\}$ indicating if $\vx_t \succ \vx_t'$, that is, if action $\vx_t$ is {\em preferred} over action $\vx'_t$. Formally, $\sP(y_t = 1 \vert \vx_t, \vx'_t) = \sP(\vx_t \succ \vx_t')$, and $y_t=0$ with probability $1-\sP(\vx_t \succ \vx_t')$.
Based on the preference history $H_t = \{(\vx_1, \vx'_1, y_1), \dots (\vx_t, \vx'_t, y_t)\}$, the agent aims to sequentially select favorable action pairs. 
Over a horizon of $T$ steps, the success of the agent is measured through the {\em cumulative dueling regret}
\begin{equation}\label{eq:dueling_regret}
    R^\mathrm{D}(T) = \sum_{t=1}^T \frac{\sP(\vx^\star \succ \vx_t) + \sP(\vx^\star \succ \vx'_t) - 1}{2},  
\end{equation}
which is the average sub-optimality gap between the chosen pair and a globally preferred action $\vx^\star$.
To better understand this notion of regret, consider the scenario where actions $\vx_t$ and $\vx_t'$ are both optimal. Then the probabilities are equal to $0.5$ and the dueling regret will not grow further, since the regret incurred at step $t$ is zero.
This formulation of $R^\mathrm{D}(T)$ is commonly used in the literature of dueling Bandits and RL with preference feedback \citep{urvoy2013generic, pacchiano2021dueling, zhu2024principled} and is adapted from \citet{yue2012k}. 
Our goal is to design an algorithm that satisfies a {\em sublinear} dueling regret, where $R^\mathrm{D}(T)/T \rightarrow 0$ as $T \rightarrow \infty$. This implies that given enough evidence, the algorithm will converge to a globally preferred action.
To this end, we take a utility-based approach and consider an unknown utility function $\tf:\calX \rightarrow\sR$, which reflects the preference via
\begin{equation}\label{eq:def_pref}
  \sP(\vx_t\succ \vx'_t)  \coloneqq \sigmoid\left( \tf(\vx_t) - \tf(\vx^\prime_t) \right)
\end{equation}
where $\sigmoid: \sR \rightarrow [0,1]$ is the sigmoid function\footnote{May be generalized to differentiable monotonically increasing functions satisfying $\sigmoid(x)+\sigmoid(-x) = 1$.}, i.e. $\sigmoid(a) = (1+e^{-a})^{-1}$. 
Referred to as the Bradley-Terry model \citep{bradley1952rank}, this probabilistic model for binary feedback is widely used in the literature for logistic and generalized bandits \citep{filippi2010parametric,faury2020improved}.
Under the utility-based model, $\vx^\star = \argmax_{\vx \in \calX}f(\vx)$ and we can draw connections to a classic bandit problem with direct feedback over the utility $f$.
In particular, \citet{saha2021optimal} shows that the dueling regret of \eqref{eq:dueling_regret} is {\em equivalent} up to constant factors, to the average {\em utility} regret of the two actions, that is $\sum_{t=1}^T \tf(\vx^\star) - [\tf(\vx_t)+ \tf(\vx_t')]/2$.\looseness=-1

Throughout this paper, we make two key assumptions over the environment. We assume that the domain $\calX \subset \sR^{d_0}$ is compact, and that the utility function lies in $\calH_k$, a Reproducing Kernel Hilbert Space corresponding to some kernel function $k \in \calX \times \calX \rightarrow \sR$ with a bounded RKHS norm $\norm{\tf}_k \leq B$. Without a loss of generality, we further suppose that the kernel function is normalized and $k(\vx, \vx)\leq 1$ everywhere in the domain.
Our set of assumptions extends the prior literature on logistic bandits and dueling bandits from linear rewards or finite action spaces, to continuous domains with non-parametric rewards.

While our theoretical framework targets euclidean domains, our methodology may be used on general domains of text or images, given vector embeddings obtained via unsupervised learning. Solving a bandit problem on top of embeddings from a pretrained language model is common practice in further fine-tuning of such models \citep[e.g.,][]{nguyen2024predicting, mehta2023sample}, and is further demonstrated in our Yelp experiment (c.f~\cref{sec:yelp}). Lastly, we highlight that our results may be smoothly extended to contextual bandits with stochastic context, by simply modifying the signature of the kernel function to $k'(\vx, \vx', \vz):  \calX\times\calX\times \calZ\rightarrow \sR$, where $\vz \in \calZ$ denotes the context. This setting accommodates applications in active learning for fine-tuning of large language models, where the context is the prompt and the two actions are two alternative responses.

\looseness=-1

%% file: sections/4_logistic_warmup.tex
\section{Kernelized Confidence Sequences with Direct Logistic Feedback}\label{sec:logistic}
As a warm-up, we consider a hypothetical scenario where $\vx'_t = \vx_{\mathrm{null}}$ for all $t\geq 1$ such that $\tf(\vx_{\mathrm{null}})=0$.
Therefore at every step, we suggest an action $\vx_t$ and receive a noisy binary feedback $y_t$, which is equal to one with probability $\sigmoid(\tf(\vx_t))$. 
This example reduces our problem to logistic bandits which has been rigorously analyzed for linear rewards \citep{filippi2010parametric, faury2020improved}. 
We extend prior work to the non-parametric setting by proposing a tractable loss function for estimating the utility function, a.k.a.~reward. We present novel confidence intervals that quantify the uncertainty on the logistic predictions {\em uniformly} over the action domain. 
In doing so, we propose confidence sequences for the kernelized logistic likelihood model that are of independent interest for developing sample-efficient solvers for online and active classification.

The feedback $y_t$ is a Bernoulli random variable, and its likelihood depends on the utility function as $s(f(\vx_t))^{y_t}[1-s(f(\vx_t))]^{1-y_t}$.
Then given history $H_t$, we can estimate $\tf$ by $\fhat$, the minimizer of the regularized negative log-likelihood loss
\begin{equation}\label{eq:logistic_loss}
    \calL_k^\mathrm{L}(f; H_t) \coloneqq \sum_{\tau=1}^t -y_\tau\log \left[\sigmoid(f(\vx_\tau))\right] - (1-y_\tau) \log \left[1-\sigmoid(f(\vx_\tau))\right] + \frac{\lambda}{2} \norm{f}^2_k
\end{equation}
where $\lambda>0$ is the regularization coefficient. 
The regularization term ensures that \smash{$\norm{\fhat}_k$} is finite and bounded. 
For simplicity, we assume throughout the main text that \smash{$\norm{ \fhat}_k \leq B$}.
However, we do not need to rely on this assumption to give theoretical guarantees. In the appendix, we present a more rigorous analysis by 
projecting $\fhat$ back into the RKHS ball of radius $B$ to ensure that the $B$-boundedness condition is met, instead of assuming it. We do not perform this projection in our experiments. \looseness-1

Solving for $\fhat$ may seem intractable at first glance since the loss is defined over functions in the large space of $\calH_k$. However, it is common knowledge that the solution has a parametric form and may be calculated by using gradient descent. This is a direct application of the Representer Theorem \citep{scholkopf2001generalized} and is detailed in \cref{prop:representer}.
\begin{proposition}[Logistic Representer Theorem]\label{prop:representer}
    The regularized negative log-likelihood loss of \eqref{eq:logistic_loss} has a unique minimizer $\fhat$, which takes the form 
$\fhat(\cdot) = \sum_{\tau=1}^t \alpha_\tau k(\cdot, \vx_\tau)$
    where $(\alpha_1, \dots \alpha_t) \eqcolon \valpha_t \in \sR^t$ is the minimizer of the strictly convex loss
    \begin{equation*}\label{eq:logistic_nll}
    \calL_k^\mathrm{L}(\valpha; H_t) = \sum_{\tau=1}^t -y_\tau\log \left[\sigmoid(\valpha^\top \vk_{t}(\vx_\tau))\right] - (1-y_\tau) \log \left[1-\sigmoid(\valpha^\top \vk_{t}(\vx_\tau))\right] + \frac{\lambda}{2} \norm{\valpha}^2_2   
    \end{equation*}
   with $\vk_t(\vx) = (k(\vx_1, \vx), \dots, k(\vx_t, \vx)) \in \sR^t$.
\end{proposition}
Given $\fhat$, we may predict the expected feedback for a point $\vx$ as $s(\fhat(\vx))$. 
Centered around this prediction, we construct confidence sets of the form $[s(\fhat(\vx))\pm \beta_t(\delta)\sigma_t(\vx)]$, and show their uniform anytime validity. The width of the sets are characterized by $\sigma_t(\vx)$ defined as
\begin{equation}\label{eq:def_sigma}
    \sigma^2_t(\vx) \coloneqq k(\vx, \vx) - \vk_t^\top(\vx) (K_t + \lambda\kappa\mI_t)^{-1}\vk_t(\vx)
\end{equation}
where $ \kappa = \sup_{a \leq B} 1/\dot{\sigmoid}(a)$, with $\dot s(a) = s(a)(1-s(a))$ denoting the derivative of the sigmoid function, and $K_t \in \sR^{t\times t}$ is the kernel matrix satisfying $[K_t]_{i,j} = k(\vx_i, \vx_j)$.
Our first main result shows that for a careful choice of $\beta_t(\delta)$, these sets contain $s(f(\vx))$ simultaneously for all $\vx \in \calX$ and $t\geq 1$ with probability greater than $1-\delta$. 
\begin{theorem}[Kernelized Logistic Confidence Sequences]\label{thm:func_CI}
 Assume $f \in \calH_k$ and $\norm{f}_k\leq B$. Let $0<\delta <1$ and set 
 \begin{equation}\label{eq:def_beta}
     \beta_t(\delta)\coloneqq 4LB + 2L\sqrt{\frac{2\kappa}{\lambda}(\gamma_t +\log 1/\delta)},
 \end{equation}
where $\gamma_t \coloneqq \max_{\vx_1, \dots, \vx_t} \tfrac{1}{2}\log\det(\mI_t + (\lambda \kappa)^{-1}K_T)$, and $L \coloneqq \sup_{a \leq B}\dot{\sigmoid}(a)$.
Then
\[
\sP\left(\forall t\geq 1, \vx\in \calX:\,  \abs{\sigmoid\left(\fhat( \vx) \right)- \sigmoid\left(\tf(\vx)\right)}\leq \beta_t(\delta)\sigma_t(\vx)\right) \geq 1-\delta.
\] 
\end{theorem}
Function-valued confidence sets around the kernelized ridge estimator are analyzed and used extensively to design bandit algorithms with noisy feedback on the true reward values \citep{valko2013finite, srinivas2009gaussian, chowdhury2017kernelized, whitehouse2023improved}. However, under noisy logistic feedback, this literature falls short as the proposed confidence sets are no longer valid for the kernelized logistic estimator $\fhat$.
One could still estimate $f$ using a kernelized ridge estimator and benefit from this line of work. However, as empirically demonstrated in \cref{fig:logit_bandit_regret}, this will not be a sample-efficient approach.

\textbf{Proof Sketch.} When minimizing the kernelized logistic loss, we do not have a closed-form solution for $\fhat$, and can only formulate it using the fact that the gradient of the loss evaluated at $\fhat$ is the null operator, i.e., $\nabla \calL(\fhat; H_t): \calH\rightarrow\calH = \bm{0}$.
The key idea of our proof is to construct confidence intervals as $\calH$-valued ellipsoids in the {\em gradient space} and show that the gradient operator evaluated at $f$ belongs to it with high probability (c.f. \cref{lem:grad_CI}). 
We then translate this back into intervals around point estimates $s(\fhat(\vx))$ uniformly for all points $\vx \in \calX$. The complete proof is deferred to \cref{app:logistic}, and builds on the results of \citet{faury2020improved} and \citet{whitehouse2023improved}.

\textbf{Logistic Bandits.} Such confidence sets are an integral tool for action selection under uncertainty, and bandit algorithms often rely on them to balance exploration against exploitation.
To demonstrate how \cref{thm:func_CI} may be used for bandit optimization with direct logistic feedback, we consider \algo, the kernelized Logistic GP-UCB algorithm. Presented in \cref{alg:kernel_log_nonparam}, it extends the optimistic algorithm of \citet{faury2020improved} from the linear to the kernelized setting, by using the confidence bound of \cref{thm:func_CI} to calculate an optimistic estimate of the reward.
We proceed to show that \algo attains a sublinear logistic regret, which is commonly defined as
\[
R^{\mathrm{L}}(T) = \sum_{i=1}^T \sigmoid(\tf(\vx^\star)) - \sigmoid(\tf(\vx_t)).
\]
To the best of our knowledge, the following corollary presents the first regret bound for logistic bandits in the kernelized setting and may be of independent interest.
\begin{corollary}\label{thm:regret_kern_logistic}
        Let $\delta \in (0, 1]$ and choose the exploration coefficients $\beta_t(\delta)$ as described in \cref{thm:func_CI} for all $t\geq 0$.
    Then \algo satisfies the anytime cumulative regret guarantee of 
\begin{align*}
\sP\left(\forall T\geq 0: R^{\mathrm{L}}(T) \leq  C_L \beta_T(\delta)\sqrt{T\gamma_t}\right) \geq 1-\delta.
\end{align*}
where $C_L \coloneqq \sqrt{8/\log (1+ (\lambda\kappa)^{-1})}$.
\end{corollary}

%% file: sections/5_main_result.tex
\section{Main Results: Bandits with Preference Feedback} \label{sec:kernelized_bandits_preference_feedback}
\looseness=-1
We return to our main problem setting in which a pair of actions, $\vx_t$ and $\vx_{t}'$, are chosen and the observed response is a noisy binary indicator of $\vx_t$ yielding a higher utility than $\vx'_t$. 
While this type of feedback is more consistent in practice, it creates quite a challenging problem compared to the logistic problem of \cref{sec:logistic}. The search space for action pairs $\calX \times \calX$ is significantly larger than $\calX$, and the observed preference feedback of $s(f(\vx_t) - f(\vx'_t))$ conveys only relative information between two actions. 
We start by presenting a solution to estimate $f$ and obtain valid confidence sets under preference feedback.
Using these confidence sets we then design the \dalgo algorithm which chooses action pairs that are not only favorable, i.e., yield high utility, but are also informative for improving the utility confidence sets.

\subsection{Preference-based Confidence Sets}
\looseness=-1
We consider the probabilistic model of $y_t$ as stated in \cref{sec:problem}, and write the corresponding regularized negative loglikelihood loss as
\begin{equation}\label{eq:preference_loss}
    \begin{aligned}
    \calL^\mathrm{D}_k(f; H_t) \coloneqq \sum_{\tau=1}^t & - y_\tau\log\left[\sigmoid(f(\vx_\tau)-f(\vx'_\tau))\right]\\
    & - (1-y_\tau)\log\left[1-\sigmoid\left(f(\vx_\tau)-f(\vx'_\tau)\right)\right]  + \tfrac{\lambda}{2} \norm{f}_k^2.
    \end{aligned}
\end{equation}
This loss may be optimized over different function classes and is commonly used for linear dueling bandits \citep[e.g.,][]{saha2021optimal}, and has been notably successful in reinforcement learning with human feedback \citep{christiano2017deep}.
We proceed to show that the preference-based loss $\calL_k^\mathrm{D}$ is equivalent to  $\calL_{\kduel}^\mathrm{L}$, the standard logistic loss \eqref{eq:logistic_loss} invoked with a specific kernel function $\kduel$. This will allow us to cast the problem of inference with preference feedback as a kernelized logistic regression problem.
To this end, we define the {\em dueling kernel} as 
\[
\kduel \big((\vx_1, \vx'_1), (\vx_2, \vx_2')\big) \coloneqq k(\vx_1, \vx_2) + k(\vx_1', \vx_2') - k(\vx_1, \vx_2') - k(\vx_1', \vx_2)
\]
for all $(\vx_1, \vx_1'), (\vx_2, \vx_2') \in \calX \times \calX$, and let $\calH_{\kduel}$ be the RKHS corresponding to it.
While the two function spaces $\calH_{\kduel}$ and $\calH_k$ are defined over different input domains, we can show that they are isomorphic, under simple regularity conditions.
\begin{proposition} \label{prop:rkhs_equival}
    Let $f:\calX\rightarrow \sR$. Consider a kernel $k$ and the sequence of its eigenfunctions $(\phi_i)_{i=1}^\infty$. 
    Assume the eigenfunctions are zero-mean, i.e. $\int_{\vx \in \calX} \phi_i(\vx)\mathrm{d}\vx = 0$. Then $f \in \calH_k$, if and only if there exists $h \in \calH_{\kduel}$ such that $h(\vx, \vx') = f(\vx) - f(\vx')$. Moreover, $\norm{h}_{\kduel} = \norm{f}_k$.
\end{proposition}
The proof is left to \cref{app:equivalence}.
The assumption on eigenfunctions in \cref{prop:rkhs_equival} is primarily made to simplify the equivalence class.
In particular, the relative preference function $h$ can only capture the utility $f$ up to a bias, i.e., if a constant bias $b$ is added to all values of $f$, the corresponding $h$ function will not change. The value of $b$ may not be recovered by drawing queries from $h$, however, this will not cause issues in terms of identifying $\argmax$ of $f$ through querying values of $h$.
Therefore, we set $b=0$ by assuming that eigenfunctions of $k$ are zero-mean.
This assumption automatically holds for all kernels that are translation or rotation invariant over symmetric domains, since their eigenfunctions are periodic $L_2(\calX)$ basis functions, e.g., Mat{\'e}rn kernels and sinusoids.

\cref{prop:rkhs_equival} allows us to re-write the preference-based loss function of \eqref{eq:preference_loss} as a logistic-type loss
\[
\calL^\mathrm{L}_{\kduel}(h; H_t) = \sum_{\tau=1}^t - y_\tau\log\left[\sigmoid(h(\vx_\tau, \vx'_\tau))\right] - (1-y_\tau)\log\left[1-\sigmoid\left(h(\vx_\tau, \vx'_\tau)\right)\right]  + \frac{\lambda}{2} \norm{h}_{k^\mathrm{D}}^2,
\]
that is equivalent to \eqref{eq:logistic_loss} up to the choice of kernel. We define the minimizer $h_t \coloneqq \argmin \calL_{\kduel}^\mathrm{L}(h; H_t)$
and use it to construct anytime valid confidence sets for the utility $f$ given only preference feedback. 
\begin{corollary}[Kernelized Preference-based Confidence Sequences]\label{cor:dueling_CI}
 Assume $f \in \calH_{k}$ and $\norm{f}_{k} \leq B$.
Choose $0<\delta <1$ and set $\beta^{\mathrm{D}}_t(\delta)$ and $\sigma^\mathrm{D}_t$ as in equations \eqref{eq:def_sigma} and \eqref{eq:def_beta}, with $k^D$ used as the kernel function. Then,
\[
 \sP\left( \forall t\geq 1, \vx, \vx' \in \calX:\, \abs{\sigmoid\left(h_t( \vx,\vx') \right)- \sigmoid\left(f(\vx) - f(\vx') \right)}\leq \beta^{\mathrm{D}}_t(\delta)\sigma^D_t(\vx, \vx')\right)\geq1-\delta.
\]    
where $h_t = \argmin \calL^\mathrm{L}_{\kduel}(h; H_t)$.
\end{corollary} 
\cref{cor:dueling_CI} gives valid confidence sets for kernelized utility functions under preference feedback and immediately improves prior results on linear dueling bandits and kernelized dueling bandits with regression-type loss, to kernelized setting with logistic-type likelihood.
To demonstrate this, in \cref{appendix:algorithm_extensions_to_kernelized}
we present the kernelized extensions of \MaxInfo (\cite{saha2021optimal}, Algorithm~\ref{alg:max_info_kernelized}), and \IDS (\cite{kirschner2021bias}, Algorithm~\ref{alg:kernel_ids}) and prove the corresponding regret guarantees (cf.~Theorems~\ref{thm:regret_maxinp}~and~\ref{thm:ids_regret}).
\cref{cor:dueling_CI} holds almost immediately by invoking \cref{thm:func_CI} with the dueling kernel $k^\mathrm{D}$ and applying \cref{prop:rkhs_equival}. A proof is provided in \cref{app:equivalence} for completeness.

\looseness=-1
\textbf{Comparison to Prior Work.} 
A line of previous work assumes that both $f$ and the probability $s(f(\vx))$ are $B$-bounded members of $\calH_k$. This allows them to directly estimate $s(f(\vx))$ via kernelized linear regression \citep{xu2020zeroth, mehta2023kernelized, kirschner2021bias}.
The resulting confidence intervals are then around the least squares estimator, which does not align with the logistic estimator $\fhat$.
This model does not encode the fact that $s(f(\vx))$ only takes values in $[0,1]$ and considers a sub-Gaussian distribution for $y_t$, instead of the Bernoulli formulation when calculating the likelihood.
Therefore, the resulting algorithms require more samples to learn an accurate reward estimate.
In a concurrent work, \citet{xu2024principled} also consider the loss function of \cref{eq:preference_loss} and present likelihood-ratio confidence sets.
The width of the sets at time $T$, scales with $\sqrt{T\log \calN(\calH_k; 1/T})$ where the second term is the {\em metric entropy} of the $B$-bounded RKHS at resolution $1/T$, that is, the log-covering number of this function class, using balls of radius $1/T$. 
It is known that $\log \calN(\calH_k; 1/T)\asymp \gamma_T$ as defined in \cref{thm:func_CI}.
This may be easily verified using \citet[Example 5.12]{wainwright2019high} and \cite[Definition 1]{vakili2021information}.
Noting the definition of $\beta_t^\mathrm{D}$, we see that likelihood ratio sets of \citet{xu2024principled} are wider than \cref{cor:dueling_CI}. Consequently, the presented regret guarantee in this work is looser by a factor of $T^{1/4}$ compared to our bound in \cref{thm:regret_maxmin}.

\begin{algorithm}[t!]
\caption{\textsc{MaxMinLCB}\label{alg:maxmin} }
\begin{algorithmic}
\State \textbf{Input}  $(\beta^\mathrm{D}_t)_{t\geq 1}$. %
    \For{$t \geq 1$}
    \State Play the most potent pair $(\vx_t, \vx'_t)$ according to the Stackelberg game
    \begin{equation*}
        \begin{aligned}
             \vx_t & = \arg \max_{\vx \in \calM_t} \sigmoid(h_t(\vx, \vx'(\vx))) - \beta^\mathrm{D}_t \sigma^\mathrm{D}_t(\vx, \vx'(\vx)) \\
            \text{s.t. } \vx'(\vx) &= \arg \min_{\vx' \in \calM_t} \sigmoid(h_t(\vx,\vx')) - \beta^\mathrm{D}_t \sigma^\mathrm{D}_t(\vx, \vx') \\
            \text{and } \vx_t' &= \vx'(\vx_t).
        \end{aligned}
    \end{equation*} 
    \State Observe $y_t$ and append history.
    \State Update $h_{t+1}$ and $\sigma^\mathrm{D}_{t+1}$ and the set of plausible maximizers
\[\calM_{t+1} = \{\vx \in \calX\vert\, \forall \vx' \in \calX:\,  \sigmoid(h_{t+1}(\vx,\vx')) + \beta^\mathrm{D}_{t+1}\sigma^\mathrm{D}_{t+1}(\vx, \vx') \geq 0.5  \}.\] 
    \EndFor
\end{algorithmic}
\end{algorithm}

\subsection{Action Selection Strategy}
Let $\vx'(\vx) = \arg \min_{\vx' \in \calM_t} \mathrm{LCB}_t(\vx, \vx')$ denote a {\em response} function. We propose \dalgo in \cref{alg:maxmin} for the preference feedback bandit problem that selects $\vx_t$ and $\vx_t'$ jointly as
\begin{equation} \label{eq:maxminlcb_objective}
    \begin{aligned}
         \vx_t & = \arg \max_{\vx \in \calM_t} \mathrm{LCB}_t(\vx, \vx'(\vx)) &&  
        \text{(Leader)} \\
        \vx'_t & = \vx'(\vx_t) && \text{(Follower)}
    \end{aligned}
\end{equation} 
where the lower-confidence bound $\mathrm{LCB}_t (\vx, \vx') = \sigmoid(h_t(\vx,\vx')) - \beta^\mathrm{D}_t \sigma^\mathrm{D}_t(\vx, \vx')$ presents a pessimistic estimate of $h$ and $\calM_t = \{\vx \in \calX\vert\, \forall \vx' \in \calX:\,  \sigmoid(h_{t}(\vx,\vx')) + \beta^\mathrm{D}_{t}\sigma^\mathrm{D}_{t}(\vx, \vx') \geq 0.5  \}$ is the set of potentially optimal actions. The second action is chosen as $\vx_t' = \vx'(\vx_t)$.
\cref{eq:maxminlcb_objective} forms a zero-sum Stackelberg (Leader--Follower) game where the actions $\vx_t$ and $\vx_t'$ are chosen sequentially \citep{stackelberg1952theory}. First, the Leader selects $\vx_t$, then the Follower selects $\vx_t'$ depending on the choice of $\vx_t$. Importantly, due to the sequential nature of action selection, $\vx_t$ is chosen by the Leader such that the Follower's action selection function, $\vx'(\cdot)$, is accounted for in the selection of $\vx_t$.
Sequential optimization problems are known to be computationally NP-hard even for linear functions \citep{jeroslow1985polynomial}. However, due to their importance in practical applications, there are algorithms that can efficiently approximate a solution over large domains \citep{sinha2017review, ghadimi2018approximation, dagreou2022framework, camacho2023metaheuristics}.\looseness-1

\looseness=-1
\dalgo builds on a simple insight: if the utility $f$ is known, both the Leader and the Follower will choose $\vx^\star$ yielding an objective value $0.5$ for both players, and zero dueling regret.
Since \dalgo has no access to $f$, it leverages the confidence sets of \cref{cor:dueling_CI} and uses a pessimistic approach by considering the LCB instead.
There are two crucial properties of the Follower specific to this game.
First, the Follower can not do worse than the Leader with respect to the $\mathrm{LCB}_t$.
In any scenario, the Follower can match the Leader's action which results in $\mathrm{LCB}_t(\vx_t, \vx_t') = 0.5$.
Second, for sufficiently tight confidence sets, the Follower will not select sub-optimal actions.
In this case, the Leader's best action must be optimal as it anticipates the Follower's response and \cref{eq:maxminlcb_objective} recovers the optimal actions.
Therefore, the objective value of the game considered in \cref{eq:maxminlcb_objective} is always less than, or equal to the objective of the game with known utility function $f$, i.e., $\mathrm{LCB}_t(\vx_t, \vx_t') \leq 0.5 = f(\vx^\star, \vx^\star)$ and the gap shrinks with the confidence sets.
Overall, the Stackelberg game in \cref{eq:maxminlcb_objective} can be considered as a lower approximation of the game played with known utility function $f$.

\looseness=-1
The primary challenge for \dalgo is to sample action pairs that sufficiently shrink the confidence sets for the optimal actions without accumulating too much regret.
\dalgo balances this exploration-exploitation trade-off naturally with its game theoretic formulation.
We view the selection of $\vx_t$ to be exploitative by trying to maximize the unknown utility $f(\vx_t)$ and minimizing regret.
On the other hand, $\vx_t'$ is chosen to be the most competitive opponent to $\vx_t$, i.e., testing whether the condition $\mathrm{LCB}_t(\vx_t, \vx_t') \geq 0.5$ holds.
Note that $\mathrm{LCB}_t$ is pessimistic concerning $\vx_t$ making it robust against the uncertainty in the confidence set estimation. At the same time, $\mathrm{LCB}_t$ is an optimistic estimate for $\vx_t'$ encouraging exploration.
In our main theoretical result, we prove that under the assumptions of \cref{cor:dueling_CI}, 
\dalgo achieves sublinear regret on the dueling bandit problem.

\begin{theorem}\label{thm:regret_maxmin}
Suppose the utility function $f$ lies in $\calH_k$ with a norm bounded by $B$, and that kernel $k$ satisfies the assumption of \cref{prop:rkhs_equival}.
        Let $\delta \in (0, 1]$ and choose the exploration coefficient 
    \smash{$\beta^{\mathrm{D}}_t(\delta)$} as in \cref{cor:dueling_CI}.
    Then \textsc{MaxMinLCB} satisfies the anytime dueling regret of 
\begin{align*}
\sP\left(\forall T\geq 0: R^{\mathrm{D}}(T) \leq C_3\beta^\mathrm{D}_T(\delta)\sqrt{T\gamma^\mathrm{D}_T} = \calO(\gamma_T^{\mathrm{D}}\sqrt{T})\right) \geq 1-\delta
\end{align*}
where $\gamma^\mathrm{D}_T$ is the $T$-step information gain of kernel $k^\mathrm{D}$ and $C_3 = (8+2\kappa)/\sqrt{\log (1+ 4(\lambda\kappa)^{-1})}$.
\end{theorem}
The proof is left to \cref{app:dueling_regret_bound}. The information gain $\gamma_T^\mathrm{D}$ in \cref{thm:regret_maxmin} quantifies the structural complexity of the RKHS corresponding to $\kduel$ and its dependence on $T$ is fairly understood for kernels commonly used in applications of bandit optimization. As an example, for a Mat{\'ern} kernel of smoothness $\nu$ defined over a $d$-dimensional domain, $\gamma_T = \tilde \calO(T^{d/(2\nu+d)})$ \citep[Remark 2, ][]{vakili2021information} and the corresponding regret bound grows sublinearly with $T$.

Restricting the optimization domain to $\calM_t \subset \calX$ is common in the literature \citep{zoghi2014relative, saha2021optimal} despite being challenging in applications with large or continuous domains.
We conjecture that \dalgo would enjoy similar regret guarantees without restricting the selection domain to $\calM_t$ as done in \cref{eq:maxminlcb_objective}. This claim is supported by our experiments in \cref{sec:experiment_dueling} which are carried out without such restriction on the optimization domain.

\begin{figure}[t]
    \label{fig:cumulative_regret}
    \centering
    \begin{subfigure}[b]{0.48\textwidth}
        \centering
        \includegraphics[width=1.0\textwidth]{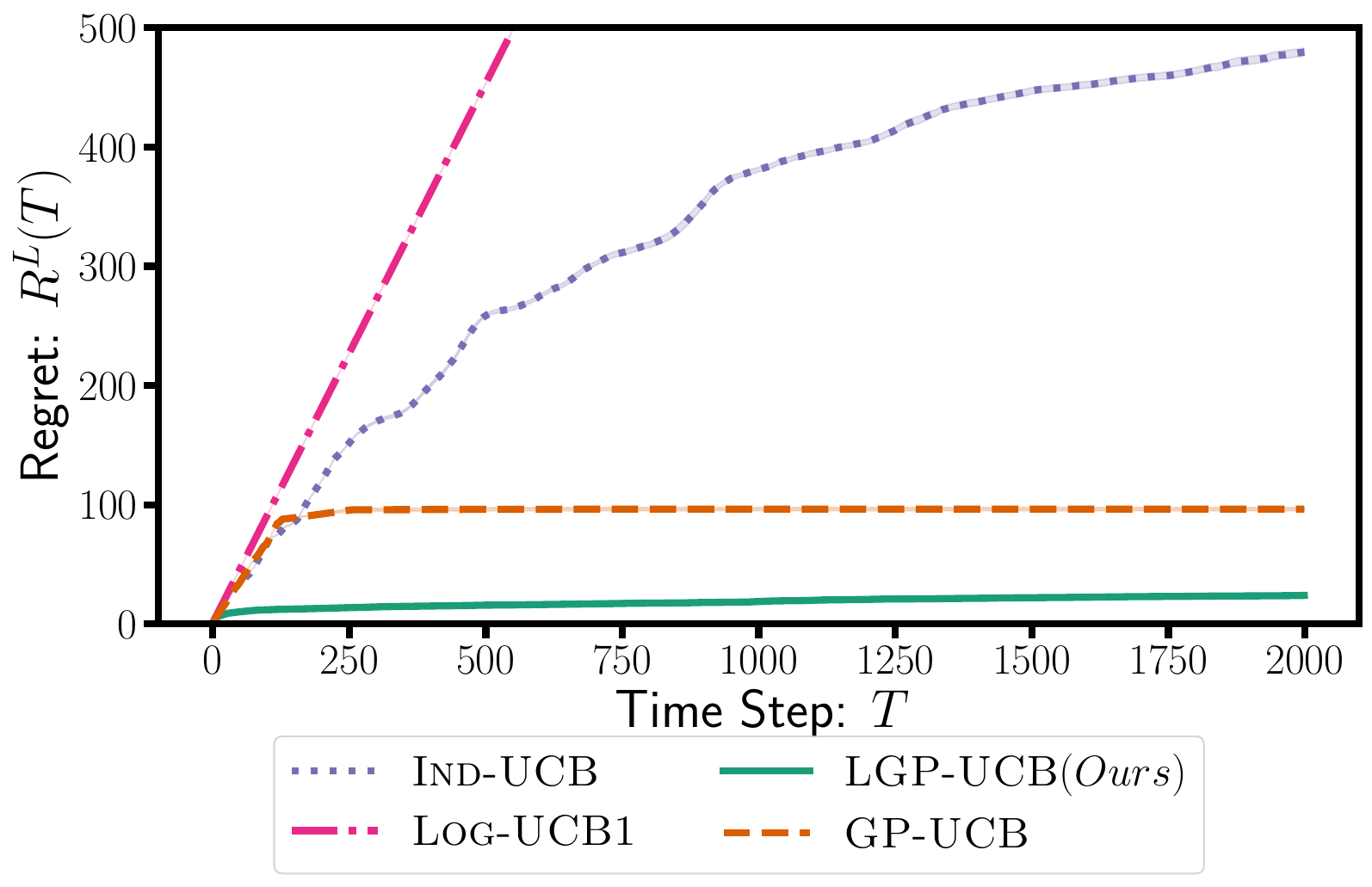}   
            \caption{Logistic feedback}
        \label{fig:logit_bandit_regret}
    \end{subfigure}
    \begin{subfigure}[b]{0.505\textwidth}
        \centering
        \includegraphics[width=1.0\textwidth]{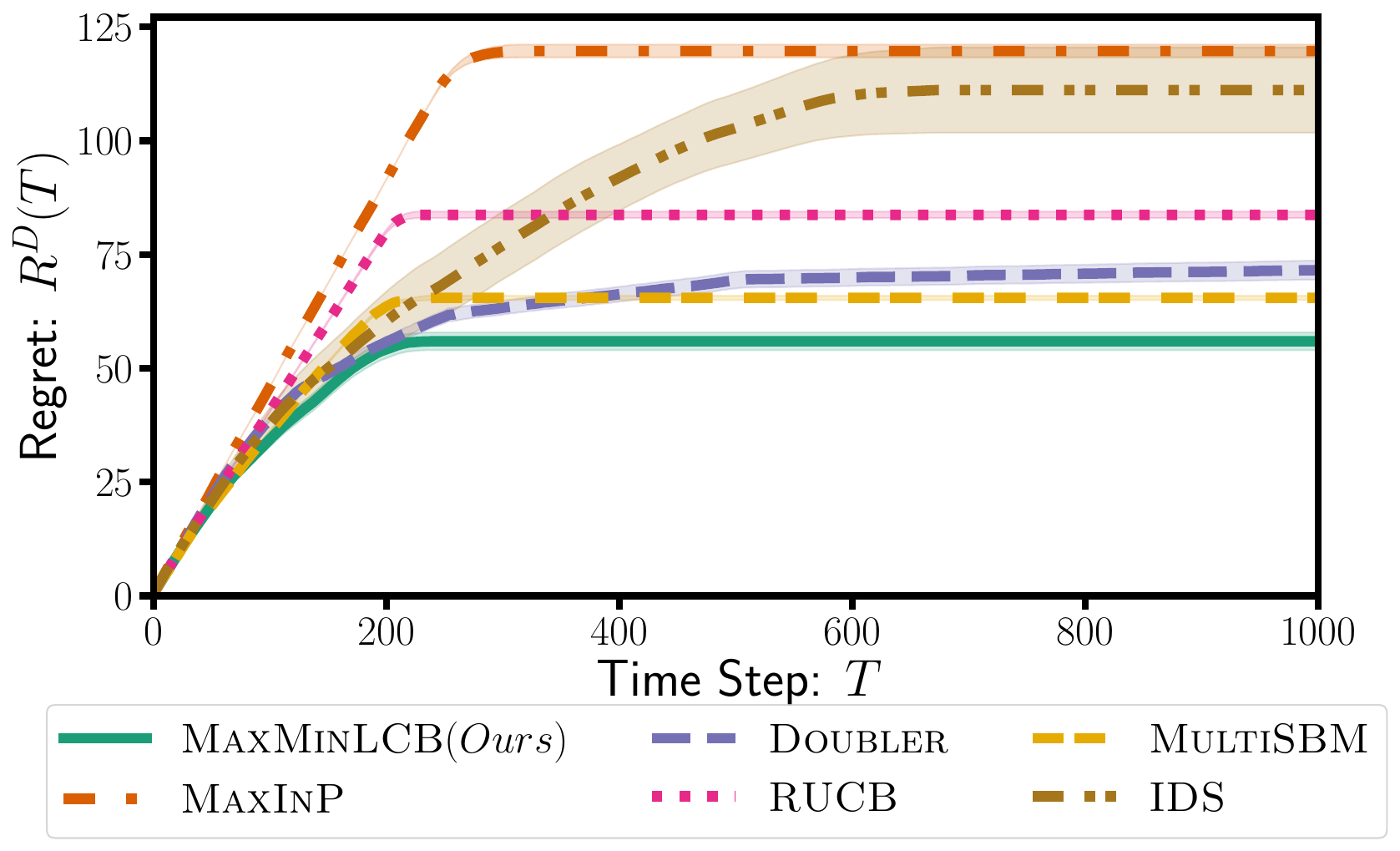}
        \caption{Preference feedback}
        \label{fig:dueling_bandit_regret}
    \end{subfigure}
    \caption{
    Regret of learning the Ackley function with logistic and preference feedback. 
    \textbf{(a)} Same UCB algorithms, each using a different confidence set. \algo performs best, showcasing the power of \cref{thm:func_CI}.
    \textbf{(b)}: Algorithms with different acquisition functions, all using our confidence sets. \dalgo is more sample-efficient.
    }
    \vspace{-10pt}
\end{figure}

%% file: sections/6_experiments.tex
\section{Experiments} \label{sec:experiments}

\looseness=-1
Our experiments are on finding the maxima of test functions commonly used in (non-convex) optimization literature \citep{jamil2013literature}, given only preference feedback. These functions cover challenging optimization landscapes including several local optima, plateaus, and valleys, allowing us to test the versatility of \dalgo.
We use the Ackley function for illustration in the main text and provide the regret plots for the remainder of the functions in \cref{appendix:additional_experiments}.
For all experiments, we set the horizon $T=2000$ and evaluate all algorithms on a uniform mesh over the input domain of size $100$.
Additionally, we conducted experiments on the Yelp restaurant review dataset to demonstrate the applicability of \dalgo on real-world data and its scaling to larger domains.
All experiments are run across $20$ random seeds and reported values are averaged over the seeds, together with standard error.
The environments and algorithms are implemented\footnote{The code is made available at \href{https://github.com/lasgroup/MaxMinLCB}{github.com/lasgroup/MaxMinLCB}.} end-to-end in JAX \citep{jax2018github}.

\subsection{Benchmarking Confidence Sets} \label{sec:logistic_bandit_experiment}
\looseness=-1
Performance of \dalgo relies on validity and tightness of the LCB. We evaluate the quality of our kernelized confidence bounds, using the potentially simpler task of bandit optimization given logistic feedback.
To this end, we fix the acquisition function for the logistic bandit algorithms to the Upper Confidence Bound (UCB) function, and benchmark different methods for calculating the confidence bound.
We refer to the algorithm instantiated with 
the confidence sets of \cref{thm:func_CI} as \algo (c.f. \cref{alg:kernel_log_nonparam}). The 
\UCB approach assumes that actions are uncorrelated, and maintains an independent confidence interval for each action as in \citet[Algorithm 3]{lattimore2020bandit}. This demonstrates how \algo utilizes the correlation between actions.
We also implement \LOGUCB \citep{faury2020improved} that assumes that $f$ is a linear function, i.e., $f(\vx) = \theta^T \vx$ to highlight the improvements gained by kernelization.
Last, we compare \algo with \GPUCB \citep{srinivas2009gaussian} that estimates probabilities $s(f(\cdot))$ via a kernelized ridge regression task. This comparison highlights the benefits of using our kernelized logistic estimator (\cref{prop:representer}) over regression-based approaches \citep{xu2020zeroth, kirschner2021bias,mehta2023kernelized,mehta2023sample}.
\cref{fig:logit_bandit_regret} shows that the cumulative regret of \algo is the lowest among the baselines.
\GPUCB performs closest to \algo, however, it accumulates regret linearly during the initial steps.
Note that \GPUCB and \algo differ in the estimation of the utility function $f_t$ while estimating the width of the confidence bounds similarly.
This result suggests that using the logistic-type loss \eqref{eq:logistic_loss} to infer the utility function is advantageous.
As expected, \UCB converges at a slower rate than \algo and \GPUCB due to ignoring the correlation between arms while \LOGUCB's regret grows linearly as the Ackley function is misspecified under the assumption of linearity.
We defer the results on the rest of the utility functions to \cref{table:logistic_comparison} in \cref{appendix:additional_experiments} and the figures therein.

\subsection{Benchmarking Acquisition Functions}
\label{sec:experiment_dueling}

\begin{table}[t]
    \caption{Benchmarking $R^{\mathrm{D}}_T$ for a variety of test utility functions, $T=2000$. The top 3 rows show results for smoother functions without steep gradients and local optima while the bottom 5 rows show the results for more challenging problems.}
    \label{table:regret_comparison}
    \centering
    \addtolength{\tabcolsep}{-0.1em}
    \begin{tabular}{ccccccc}
    \toprule
    $f$ & \dalgo & \Doubler & \MultiSBM & \MaxInfo & \RUCB & \IDS \\
    \midrule
    Branin & $\boldsymbol{104} \pm 13$ & $114 \pm 9$ & $\boldsymbol{89} \pm 13$ & $340 \pm 2$ & $\boldsymbol{101} \pm 14$ & $163 \pm 22$ \\
    Matyas & $125 \pm 5$ & $136 \pm 4$ & $\boldsymbol{106} \pm 7$ & $136 \pm 6$ & $\boldsymbol{106} \pm 6$ & $128 \pm 5$ \\
    Rosenbrock & $\boldsymbol{27} \pm 4$ & $44 \pm 12$ & $\boldsymbol{25} \pm 5$ & $109 \pm 2$ & $58 \pm 7$ & $58 \pm 13$ \\
    \hline
    Ackley & $\boldsymbol{56} \pm 2$ & $72 \pm 2$ & $65 \pm 0.5$ & $120 \pm 1$ & $84 \pm 0.7$ & $111 \pm 9$ \\
    Eggholder & $\boldsymbol{113} \pm 6$ & $154 \pm 4$ & $134 \pm 3$ & $230 \pm 34$ & $213 \pm 40$ & $141 \pm 12$ \\
    Hoelder & $\boldsymbol{141} \pm 26$ & $\boldsymbol{154} \pm 3$ & $\boldsymbol{136} \pm 15$ & $204 \pm 20$ & $200 \pm 28$ & $\boldsymbol{132} \pm 15$ \\
    Michalewicz & $\boldsymbol{138} \pm 14$ & $183 \pm 11$ & $\boldsymbol{155} \pm 10$ & $260 \pm 40$ & $269 \pm 46$ & $188 \pm 21$ \\
    Yelp & $\boldsymbol{175} \pm 22$ & $263 \pm 28$ & $\boldsymbol{199} \pm 25$ & $409 \pm 15$ & $214 \pm 22$ & $255 \pm 22$ \\
    \bottomrule
    \end{tabular}
    \vspace{-10pt}
\end{table}

\looseness=-1
In this section, we compare \dalgo with other utility-based bandit algorithms. 
To isolate the benefits of our acquisition function, we instantiate all algorithms with the same confidence sets, and use our improved preferred-based bound of \cref{cor:dueling_CI}.
Therefore, our implementation differs from the corresponding references, while we refer to the algorithms by their original name.
We consider the following baselines.
\Doubler and \MultiSBM \citep{ailon2014reducing} choose $\vx_t$ as a {\em reference} action from the recent history of actions and pair it with $\vx'_t$ which maximizes the joint UCB (cf.~\cref{alg:Doubler} and~\ref{alg:MultiSBM}).
\RUCB \citep{zoghi2014relative} chooses $\vx'_t$ similarly, however, it selects the reference action uniformly at random from $\calM_t$ (\cref{alg:RUCB}).
\MaxInfo \citep{saha2021optimal} also maintains the set of plausible maximizers $\calM_t$, and at each time step, it selects the pair of actions that maximize $\sigma_t^D(\vx, \vx')$ (\cref{alg:max_info_kernelized}).
\IDS \citep{kirschner2021bias} selects the reference action greedily by maximizing $f_t$, and pairs it with an informative action (\cref{alg:kernel_ids}). 
Notably, all algorithms, with the exception of \MaxInfo, choose one of the actions independently and use it as a reference point when selecting the other one.
\cref{fig:action_selection} illustrates the differences in action selection between UCB, maximum information, and \dalgo approaches.
We note that \POPBO \citep{xu2024principled} and \MultiSBM only differ in the estimation of the confidence set. Since we deploy the same confidence set for all acquisition functions, the two algorithms are equivalent and we use \MultiSBM in our results, however, comparisons hold for \POPBO as well.

\looseness=-1
\cref{fig:dueling_bandit_regret} benchmarks the algorithms using the Ackley utility function, where \dalgo outperforms the baselines.
All algorithms suffer from close-to-linear regret during the initial stages of learning, suggesting that there is an inevitable exploration phase. 
Notably, \dalgo, \IDS, and \Doubler are the first to select actions with high utility, while \RUCB and \MaxInfo explore for longer.
\cref{table:regret_comparison} shows the dueling regret for all utility functions.
\dalgo consistently outperforms the baselines across the analyzed functions and achieves a low standard error, supporting its efficiency in balancing exploration and exploitation in the preference feedback setting.
Only \MultiSBM shows comparable performance to \dalgo and even outperforms it on the Matyas function which is a relatively smooth function posing a simple optimization problem. However, \dalgo achieves lower regret on the Ackley and Eggholder functions which obtain many local optima and steeper gradients showing that \dalgo is preferable for challenging optimization problems.
Other acquisition functions work well only in certain cases, e.g., \IDS achieves the smallest regret for optimizing Matyas, while \RUCB excels on the Branin function.
This indicates the challenges each utility function offers and the performance of the action selection is task dependent.
The consistent performance of \dalgo demonstrates its robustness against the underlying unknown utility function.

\subsection{Real-world Experiment}
\label{sec:yelp}
To demonstrate the scalability and applicability of \dalgo, we conduct an experiment on the Yelp dataset of restaurant reviews, which consists of $275$ restaurants and $20$ users after the pre-processing stage.
For each user, we want to find the restaurants that best fit their preference, via making sequential recommendations and receiving comparative feedback.
We define the action space $\calX$ by assigning to each restaurant their respective $32$-dimensional embedding of their reviews, i.e., $\calX \subseteq \sR^{32}$.
The dataset provides utility values for users in the form of ratings on the scale of $1$ to $5$, however, not all users rated every restaurant.

\begin{figure}[t]
    \centering
    \includegraphics[width=0.55\textwidth]{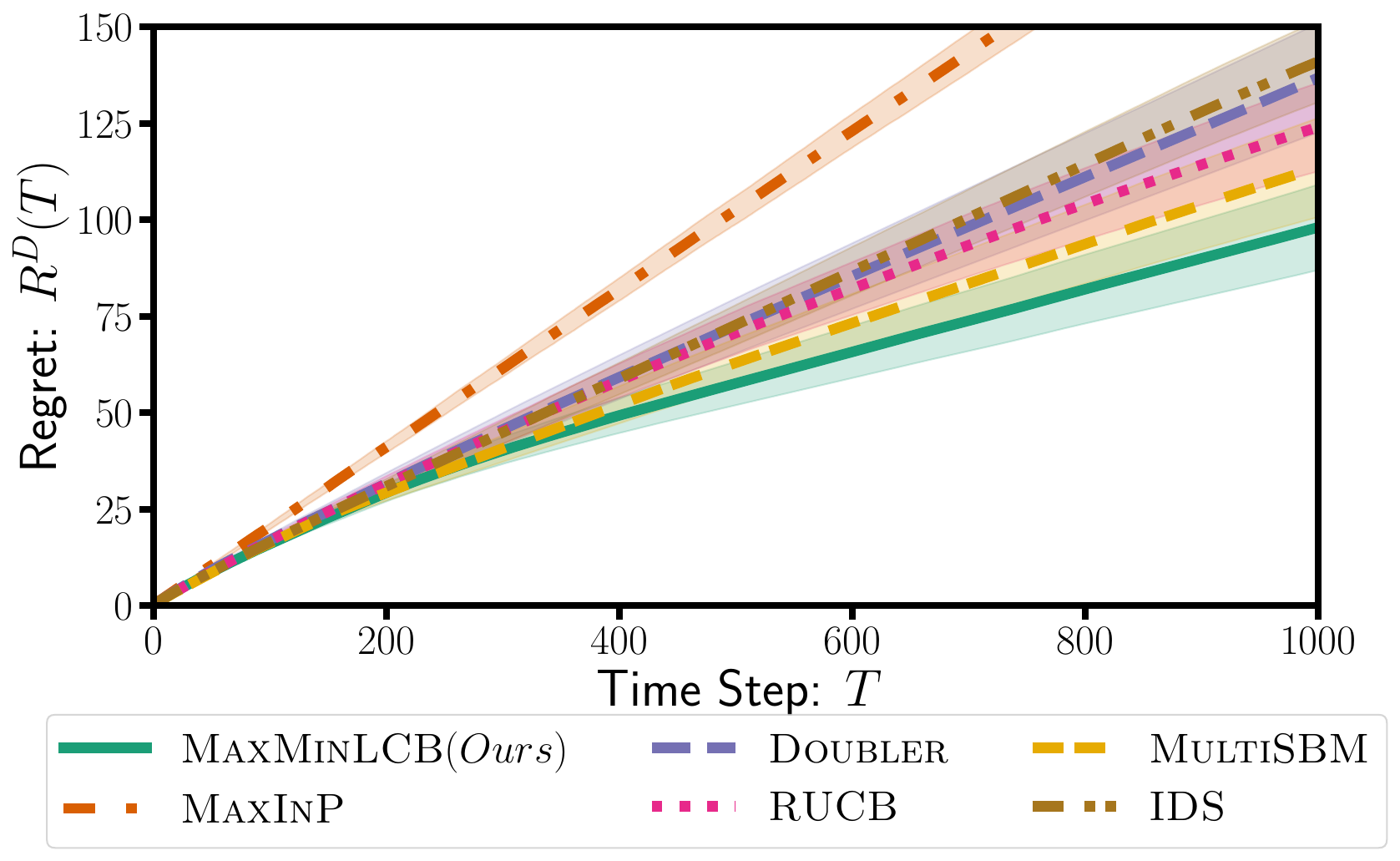}
    \caption{
    \algo is more sample-efficient when making restaurant recommendations based on Yelp open dataset with preference feedback. All baselines use the confidence sets of Cor.~\ref{cor:dueling_CI}.
    }
    \label{fig:yelp_regret}
    \vspace{-10pt}
\end{figure}

We estimate missing ratings using collaborative filtering \citep{schafer2007collaborative}.
Further details on the data processing are deferred to \cref{appendix:yelp_experiments}.
\looseness-1 \cref{fig:yelp_regret} shows that the results of this larger problem align with previous conclusions. \dalgo remains the best-performing algorithm with \MultiSBM following second. The cumulative regret is also reported in \cref{table:regret_comparison}.
Note that neither of the algorithms is tuned or modified for this experiment. These results are only intended to demonstrate that 1) the computations easily scale, and 2) the kernelized approach is still applicable in a text-based domain, by using high-quality vector embeddings.
\looseness=-1

%% file: sections/7_conclusion.tex
\section{Conclusion}

\looseness=-1
We addressed the problem of bandit optimization with preference feedback over large domains and complex targets.
We proposed \dalgo, which takes a game-theoretic approach to the problem of action selection under comparative feedback, and naturally balances exploration and exploitation by constructing a zero-sum Stackelberg game between the action pairs.
\dalgo achieves a sublinear regret for kernelized utilities, and performs competitively across a range of experiments.
Lastly, by uncovering the equivalence of learning with logistic or comparative feedback, we propose kernelized preference-based confidence sets, which may be employed in adjacent problems, such as reinforcement learning with human feedback.
The technical setup considered in this work serves as a foundation for a number of applications in mechanism design, such as preference elicitation and welfare optimization from multiple feedback sources for social choice theory, which we leave as future work.

%% file: appendices/1_logistic_proofs.tex
\section{Proofs for Bandits with Logistic Feedback}\label{app:logistic}

We have written the equations in the main text in terms of kernel matrices and function evaluations, for easier readability. For the purpose of the proof however, we mainly rely on entities in the Hilbert space. Consider the operator $\vphi: \calX \rightarrow \calH$ which corresponds to kernel $k$ and satisfies $k(\vx, \cdot) = \vphi(\vx)$. 
Then by Mercer's theorem, any \smash{$f \in \calH_k$} may be written as $f = \vtheta^\top \vphi$, where $\vtheta \in \ell_2(\sN)$ and has a $B$-bounded $\ell_2$ norm.
For a sequence of points $\vx_1, \dots, \vx_t \in \calX$, we define the (infinite-dimensional) feature map 
\smash{$\Phi_t = \left[ \vphi(\vx_1), \cdots, \vphi(\vx_t)\right]^\top$}, which gives rise to the kernel matrix $K_t: \sR^{t}\rightarrow \sR^t$ and the covariance operator $S_t: \calH_k \rightarrow \calH_k$, respectively defined as $K_t = \Phi_t \Phi_t^\top$ and $S_t = \Phi_t^\top \Phi_t$.
Let $\mI_t$ denote the $t$-dimensional identity matrix, and $\mI_\calH$ be the identity operator acting on the RKHS. Then it is widely known that the covariance and kernel operators are connected via $\det (\mI_\calH + \rho^{-2} S_t) = \det (\mI_t + \rho^{-2} K_t)$ for any $t\geq 1$ and $\rho \neq 0$.
For operators on the Hilbert space, $\det (A)$ refer to a Fredholm determinant \citep[c.f. ][]{lax2002functional}.
Lastly, in the appendix, we refer to the true unknown utility function as $f^\star(\vx) = \vphi^\top(\vx)\vtheta^\star$. In the main text, the true utility is simply referred to as $f$.
\looseness -1

To analyze our function-valued confidence sequences, we start by re-writing the logistic loss function
\begin{align*}
     \calL(\vtheta; H_t) = & \sum_{\tau=1}^t - y_\tau\log\sigmoid\left(\vtheta^\top\vphi(\vx_\tau)\right) - \sum_{\tau=1}^t  (1-y_\tau)\log\left(1-\sigmoid\left(\vtheta^\top \vphi(\vx_\tau) \right)\right)  + \frac{\lambda}{2} \norm{\vtheta}_2^2
\end{align*}
which is strictly convex in $\vtheta$ and has a unique minimizer $\vtheta_t$ which satisfies
\[
\nabla \calL(\vtheta_t; H_t) = \sum_{\tau=1}^t -y_\tau \vphi(\vx_\tau) +  g_t(\vtheta_t) = 0
\]
where $g_t(\vtheta): \calH \rightarrow \calH$ is a linear operator defined as
\[
g_t(\vtheta) \coloneqq \sum_{\tau=1}^t \vphi(\vx_\tau)
 \sigmoid(\vtheta^\top\vphi(\vx_\tau)) + \lambda \vtheta.
\]

In the main text, we assumed that minimizer of $\calL$ satisfies the norm boundedness condition. Here, we present a more rigorous analysis which does not assume so. Instead, we work with a projected estimator defined via
\begin{align}
    \vtheta_t^P & = \argmin_{\norm{\vtheta}_2 \leq B} \norm{\vg_t(\vtheta) - \vg_t(\vtheta_t)}_{V^{-1}_{t}}.
    \label{eq:def_thetap}
\end{align}
 where \smash{$V_t  = S_t + \kappa \lambda \mI_\calH$} and $\vtheta_t$ is the minimizer of $\calL(\vtheta; H_t)$.
Roughly put, $\vtheta_t^P \in \ell_2(\sN)$ is a norm $B$-bounded alternative to $\vtheta_t$, who also satisfies a small $\nabla \calL$, and therefore, is expected to result in an accurate decision boundary. 
We will present our proof in terms of $\vtheta_t^P$. This also proves the results in the main text, since $\vtheta_t^P = \vtheta_t$ if $\vtheta_t$ itself happens to have a $B$-bounded norm, as assumed in the main text. \looseness-1

Our analysis relies on a concentration bound for $\calH$-valued martingale sequences stated in \cite{abbasi2013online} and later in \cite{whitehouse2023improved}. Below, we have adapted the statement to match our notation.\looseness-1
\begin{lemma}[Corollary 1 \cite{whitehouse2023improved}]\label{lem:whitehouse}
Suppose the sequence $(\vx_t)_{t\geq 1}$ is $(\calF_{t})_{t\geq 1}$-adapted, where $\calF_{t}\coloneqq \sigma\left(\vx_1, \dots, \vx_t, \varepsilon_1, \dots, \varepsilon_{t-1}\right)$ and $\varepsilon_t$ are i.i.d.~zero-mean $\sigma$-subGaussian noise. Consider the RKHS $\calH$ corresponding to a kernel $k(\vx, \vx') = \vphi^\top(\vx)\vphi(\vx')$. 
 Then, for any $\rho > 0$ and $\delta \in (0, 1)$, we have that, with probability at least $1 - \delta$, simultaneously for all $t \geq 0$,
\begin{align*}
\norm{\sum_{\tau\leq t}\varepsilon_\tau\vphi(\vx_\tau)}_{V_t^{-1}} \leq \sigma \sqrt{2\log\left(\frac{1}{\delta}\sqrt{\det(\mI_t +\rho^{-2}K_t)}\right)}
\end{align*}
where $V_t  = S_t + \rho^2 \mI_\calH$.
\end{lemma}

The following lemma, which extends \citet[Lemma 8] {faury2020improved} to $\calH$-valued operators, expresses the closeness of $\vtheta_t$ and $\vtheta^\star$ in the gradient space, with respect to the norm of the covariance matrix.
\begin{lemma}[Gradient Space Confidence Bounds]\label{lem:grad_CI}
Set $0<\delta<1$. Under the assumptions of \cref{thm:func_CI}
    \[
    \sP\left(\forall t\geq 0: \, \norm{\vg_t(\vtheta_t) - \vg_t(\vtheta^\star)}_{V_t^{-1}} \leq \frac{1}{2}  \sqrt{2\log 1/\delta + 2\gamma_T}  + \sqrt{\frac{\lambda}{\kappa}} B\right) \geq 1-\delta
    \]
    where $V_t  = S_t + \kappa \lambda \mI_\calH$.
\end{lemma}
\begin{proof}[Proof of \cref{lem:grad_CI}]
    Recall that \smash{$\vg_t(\vtheta) =  \sum_{\tau\leq t} \sigmoid(\vtheta^\top \vphi(\vx_\tau)) \vphi(\vx_\tau) + \lambda \vtheta$}. Then it is straighforward to show that
    \[
    \nabla \calL(\vtheta; H_t) = \sum_{\tau \leq t} y_\tau \vphi(\vx_\tau) - g_t(\vtheta).
    \]
    Since $\vtheta_t$ is a minimizer of $\calL_t$, it holds that \smash{$\vg_t(\vtheta_t) = \sum_{\tau \leq t} y_\tau \vphi(\vx_\tau)$}. This allows us to write,
    \begin{align}
        \norm{\vg_t(\vtheta_t) - \vg_t(\vtheta^\star)}_{V_t^{-1}} & =  \norm{ \sum_{\tau \leq t} \left( y_\tau - \sigmoid(\vphi^\top(\vx_\tau)\vtheta^\star)\right) \vphi(\vx_\tau) - \lambda \vtheta^\star}_{V_t^{-1}}\notag\\
        & \leq \norm{\sum_{\tau \leq t}\varepsilon_\tau\vphi(\vx_\tau)}_{V_t^{-1}} + \lambda \norm{\vtheta^\star}_{V_t^{-1}}\label{eq:grad_con_decompose}
    \end{align}
    where $\varepsilon_\tau = y_\tau - \sigmoid(\vphi^\top(\vx_\tau)\vtheta^\star) $ is a history dependent random variable in $[0,1]$ due to the data model.
    To bound the first term, we recognize that any random variable in $[0,1]$ is $\sigma$ sub-Gaussian with $\sigma = 0.5$ and apply \cref{lem:whitehouse}. We obtain that for all $t\geq 0$, with probability greater than $1-\delta$
    \begin{align*}
\norm{\sum_{\tau \leq t}\varepsilon_\tau\vphi(\vx_\tau)}_{V_t^{-1}} &\leq \frac{1}{2} \sqrt{2\log\left(\frac{1}{\delta}\sqrt{\det(\mI_t +(\lambda\kappa)^{-1}K_t)}\right)}\\
& \leq \frac{1}{2} \sqrt{2\log 1/\delta + 2\gamma_T} 
\end{align*}
since \smash{$\gamma_t(\rho) = \sup_{\vx_1, \dots, \vx_t} \tfrac{1}{2} \log\det(\mI_t +\rho^{-2}K_t))$}.
    To bound the second term in \eqref{eq:grad_con_decompose}, note that $S_t = \Phi_t^\top \Phi_t$ is PSD and therefore $V_t \geq \kappa \lambda \mI_\calH$. Then
    \begin{equation*}
        \lambda \norm{\vtheta^\star}_{V_t^{-1}} \leq \frac{\lambda}{\sqrt{\lambda\kappa}} \norm{\vtheta^\star}_2 \leq \sqrt{\frac{\lambda}{\kappa}} B.
    \end{equation*}
    concluding the proof.
\end{proof}

The following lemma shows the validity of our parameter-space confidence sets.
\begin{lemma}\label{lem:CS_valid} Set $0<\delta<1$ and consider the confidence sets
\begin{align*}
    \Theta_t(\delta) & \coloneqq \left\{\norm{\vtheta}\leq B, \norm{\vtheta - \vtheta_t^P}_{V_t} \leq 2\sqrt{\lambda\kappa}B + \kappa \sqrt{2\log 1/\delta + 2\gamma_T} \right\}.
\end{align*}
Then,
\[
\sP\left( \forall t\geq 0:\, \vtheta^\star \in \Theta_t(\delta)\right) \geq 1-\delta
\]
\end{lemma}
\begin{proof}[Proof of \cref{lem:CS_valid}]
    We check if $\vtheta^\star \in \Theta_t(\delta)$ by bounding the following norm 
    \begin{align*}
        \norm{\vtheta^\star -\vtheta_t^P}_{V_t} & \leq \kappa \norm{ g_t(\vtheta^\star) - g_t(\vtheta^P_t) }_{V_t^{-1}} \tag*{\text{(Lem.~\ref{lem:g_manip})}}\\
        & \leq \kappa \left(\norm{ g_t(\vtheta^\star) - g_t(\vtheta_t) }_{V_t^{-1}} + \norm{ g_t(\vtheta_t) - g_t(\vtheta^P_t) }_{V_t^{-1}} \right)\\
        & \leq 2\kappa \norm{ g_t(\vtheta^\star) - g_t(\vtheta_t) }_{V_t^{-1}} \tag*{\text{(Eq.~\ref{eq:def_thetap})}}\\
        & \leq \kappa \sqrt{2\log 1/\delta + 2\gamma_T}  + 2\sqrt{\lambda \kappa} B \tag*{\text{(Lem.~\ref{lem:grad_CI})}}
    \end{align*}
\end{proof}

Lastly, we prove a formal extension of \cref{thm:func_CI}.

\begin{theorem}[\cref{thm:func_CI} - Formal]\label{lem:func_CI}

Set $0<\delta<1$ and consider the confidence sets $\calE_t(\delta) \subset \calH_k$
\[
\calE_t(\delta) = \left\{f(\cdot) = \vtheta^\top\vphi(\cdot):\, \vtheta \in \Theta_t(\delta)\right\}.
\] %
Then, simultanously for all $\vx \in \calX$, $f \in \calE_t(\delta)$ and $t\geq 0$
\[
\abs{\sigmoid(f(\vx)) - \sigmoid(f^\star(\vx))} \leq \beta_t(\delta)\sigma_t(\vx)
\]    
with probability greater than $1-\delta$, where
\[
\beta_t(\delta) \coloneqq 4LB + 2L\sqrt{\frac{\kappa}{\lambda}}\sqrt{2\log 1/\delta + 2\gamma_T}
\]
\end{theorem}
\begin{proof}[Proof of \cref{lem:func_CI}]
For simplicity in notation let us define 
\[
\tilde \beta_t(\delta) \coloneqq 2\sqrt{\lambda\kappa}B + \kappa \sqrt{2\log 1/\delta + 2\gamma_t}.
\]
Suppose $f = \vtheta^\top\vphi(\cdot)$ is in $\calE_t(\delta)$. Then
    \begin{align*}
       \abs{ \sigmoid(\vphi^\top(\vx)\vtheta^\star) - \sigmoid(\vphi^\top(\vx)\vtheta)} & = \abs{\alpha(\vx; \vtheta, \vtheta^\star) \vphi^\top(\vx) (\vtheta - \vtheta^\star)}\tag*{\text{Lem.~\ref{lem:mean_value}}}\\
       & \leq L\abs{\vphi^\top(\vx) (\vtheta - \vtheta^\star)}\tag*{\text{$\sigmoid$ is $L$-Lipschitz}}\\
       & \leq L \norm{\vphi(\vx)}_{V_t^{-1}} \norm{\vtheta-\vtheta^\star}_{V_t}\\
       & \leq L \norm{\vphi(\vx)}_{V_t^{-1}} \left( \norm{\vtheta-\vtheta^P_t}_{V_t} + \norm{\vtheta^P_t-\vtheta^\star}_{V_t} \right)\\
       & \leq L\norm{\vphi(\vx)}_{V_t^{-1}} \left( \tilde \beta_t(\delta) + \norm{\vtheta_t^P-\vtheta^\star}_{V_t} \right) \tag*{\text{$\vtheta \in \Theta_t(\delta)$}}\\
       & \stackrel{\text{w.h.p.}}{\leq} 2L\tilde \beta_t(\delta)\norm{\vphi(\vx)}_{V_t^{-1}}\tag*{\text{Lem.~\ref{lem:CS_valid}}} \\
       & \leq \frac{2L\tilde \beta_t(\delta)}{\sqrt{\lambda\kappa}}\sigma_t(\vx)\tag*{\text{Lem.~\ref{lem:norm_conversion}}}\\
       & =   \sigma_t(\vx)\left(4LB + 2L\sqrt{\frac{\kappa}{\lambda}}\sqrt{2\log 1/\delta + 2\gamma_T}\right)
    \end{align*}
    where the third to last inequality holds with probability greater than $1-\delta$ , but the rest of the inequalities hold deterministically.
\end{proof}
\begin{algorithm}[t]
\caption{\algo\label{alg:kernel_log_nonparam} }
\begin{algorithmic}
\State \textbf{Initialize} Set $(\beta_t)_{t\geq 1}$ according to \cref{thm:func_CI}.
\For{$t \geq 1$}
\State Choose an optimistic action via
\[
x_{t} = \argmax_{x \in \calX} \sigmoid(f_{t-1}(\vx)) + \beta_{t-1}(\delta)\sigma_{t-1}(\vx)
\]
\State Observe $y_t$ and append history.
\State Calculate $\fhat$ acc. to \cref{prop:representer} and update $\sigma_t$ acc. to \cref{thm:func_CI}.
\EndFor
\end{algorithmic}
\end{algorithm}
Given the confidence set of \cref{thm:func_CI}, we give extend the \textsc{LGP-UCB} algorithm of \citeauthor{faury2020improved} to the kernelized setting (c.f. \cref{alg:kernel_log_nonparam}) and prove that it satisfies sublinear regret.
\begin{proof}[Proof of \cref{thm:regret_kern_logistic}]
    Recall that if $\vx_t$ is the maximizer of the UCB, then
    \[
     \sigmoid(\vphi^\top(\vx^\star)\vtheta^P_t)  - \sigmoid(\vphi^\top(\vx_t)\vtheta^P_t) \leq   \sigma_t(\vx_t)\beta_t(\delta) - \sigma_t(\vx^\star)\beta_t(\delta) 
    \]
    Then using \cref{lem:func_CI}, we obtain the following for the regret at step $t$,
    \begin{align*}
        r_t & = \sigmoid(\vphi^\top(\vx^\star)\vtheta^\star) - \sigmoid(\vphi^\top(\vx_t)\vtheta^\star)\\
         & = \sigmoid(\vphi^\top(\vx^\star)\vtheta^\star) - \sigmoid(\vphi^\top(\vx^\star)\vtheta^P_t)  +  \sigmoid(\vphi^\top(\vx_t)\vtheta^P_t)- \sigmoid(\vphi^\top(\vx_t)\vtheta^\star)\\
         &\qquad\qquad + \sigmoid(\vphi^\top(\vx^\star)\vtheta^P_t) - \sigmoid(\vphi^\top(\vx_t)\vtheta^P_t) \\
         & \leq  \sigma_t(\vx^\star)\beta_t(\delta) +  \sigma_t(\vx_t)\beta_t(\delta) + \sigma_t(\vx_t)\beta_t(\delta) - \sigma_t(\vx^\star)\beta_t(\delta)\\
        & \leq 2 \beta_t(\delta)  \sigma_t(\vx_t)
    \end{align*}
    with probability greater than $1-\delta$ for all $t\geq 0$.
    Which allows us to bound the cumulative regret as,
    \begin{align*}
        R_T  = \sum_{t=1}^T r_t &\leq \sqrt{T\sum_{t=1}^T r_t^2} \\
        & \leq 2\beta_T(\delta) \sqrt{T \sum_{t=1}^T \sigma^2_t(\vx_t)} \tag*{\text{$\beta_t(\delta) \leq \beta_T(\delta)$}}\\
        & \leq C_1\beta_T(\delta)\sqrt{T\gamma_t} \tag*{\text{Lem.~\ref{lem:basic_infogainsum}}}
    \end{align*}
    where $C_1 \coloneqq \sqrt{8/\log (1+ (\lambda\kappa)^{-1})}$.
\end{proof}

\section{Helper Lemmas for \texorpdfstring{\cref{app:logistic}}{}}
\begin{lemma}[Mean-Value Theorem]\label{lem:mean_value}
For any $\vx \in \calX$ and $\vtheta_1, \vtheta_2 \in \ell_2(\sN)$ it holds that
\[
\sigmoid(\vtheta_2^\top\vphi(\vx)) - \sigmoid(\vtheta_1^\top\vphi(\vx)) = \alpha(\vx; \vtheta_1, \vtheta_2) (\vtheta_2-\vtheta_1)^\top \vphi(\vx)
\]
where
    \begin{equation}
        \label{eq:def_alpha_operator}
            \alpha(\vx; \vtheta_1, \vtheta_2) = \int_0^1 \dot\sigmoid(\nu \vtheta_2^\top \vphi(\vx) + (1-\nu)\vtheta_1^\top\vphi(\vx))\mathrm{d}\nu
    \end{equation}
\end{lemma}
\begin{proof}[Proof of \cref{lem:mean_value}]
For any differentiable function $\sigmoid: \sR\rightarrow \sR$ by the fundamental theorem of calculus we have
\[
\sigmoid(z_2) - \sigmoid(z_1) = \int_{z_1}^{z_2} \dot\sigmoid(z)\mathrm{d}z.
\]
We change the variable of integration to $\nu = (z-z_1)/(z_2-z_1)$, then $z = \nu z_2 + (1-\nu)z_1$ and re-writing the integral in terms of $\nu$ gives,
\[
\sigmoid(z_2) - \sigmoid(z_1) = (z_2 - z_1) \int_0^1 \dot\sigmoid(\nu z_2 + (1-\nu)z_1) \mathrm{d}\nu.
\]
Letting $z_1 = \vtheta_1^\top \vphi(\vx)$ and $z_2 = \vtheta_2^\top \vphi(\vx)$ concludes the proof.
\end{proof}
\begin{lemma}[Gradients to Parameters Conversion]\label{lem:g_manip}
For all $t\geq 0$ and norm $B$-bounded $\vtheta_1, \vtheta_2 \in \ell_2(\sN)$
    \[
    \norm{\vtheta_1 -\vtheta_2}_{V_t} \leq  \kappa \norm{ g_t(\vtheta_1) - g_t(\vtheta_2) }_{V_t^{-1}}.
    \]
\end{lemma}
\begin{proof}[Proof of \cref{lem:g_manip}]
We prove the lemma through an auxiliary operator $G_t(\vtheta_1, \vtheta_2)$ operating on $\calH_k$ \looseness-1
    \begin{align*}
        G_t(\vtheta_1, \vtheta_2) = \lambda \mI_\calH + \sum_{\tau \leq t}\alpha(\vx_\tau; \vtheta_1, \vtheta_2)\vphi(\vx_\tau)\vphi^\top(\vx_\tau)
    \end{align*}
   where $\alpha$ is defined in \eqref{eq:def_alpha_operator}.
   
   \textbf{Step 1.} First we establish how we can go back and forth between the operator norms defined based on $G_t$ and $V_t$.
    Recall that $\kappa = \sup_{z\leq B} \tfrac{1}{\dot\sigmoid(z)}$. Therefore, $\kappa^{-1} \leq \dot\sigmoid(z)$ for all $z < B$,
    implying that $\alpha(\vx; \vtheta_1, \vtheta_2) \geq \int_0^1 \kappa^{-1}\mathrm{d}\nu = \kappa^{-1}$. We can then conclude,
    \begin{equation}\label{eq:Gt_Vt}
        G_t(\vtheta_1, \vtheta_2) \geq \lambda \mI_\calH + \sum_{\tau \leq t}\kappa^{-1}\vphi(\vx_\tau)\vphi^\top(\vx_\tau) = \kappa^{-1}V_t.
    \end{equation}
   
    \textbf{Step 2.} Now by the definition of $g_t(\vtheta)$,
    \begin{align*}
        g_t(\vtheta_2) - g_t(\vtheta_1)  & =\lambda(\vtheta_2 - \vtheta_1) + \sum_{\tau \leq t}\vphi(\vx_\tau)  \left[ \sigmoid(\vtheta_2^\top\vphi(\vx_\tau)) - \sigmoid(\vtheta_1^\top\vphi(\vx_\tau)) \right]\\
        & = \lambda(\vtheta_2 - \vtheta_1) + \sum_{\tau \leq t} \vphi(\vx_\tau) \left[\alpha(\vx_\tau; \vtheta_1, \vtheta_2) \vphi^\top (\vx_\tau)(\vtheta_2-\vtheta_1)\right] \tag{\text{Lem.~\ref{lem:mean_value}}}\\
        &=  \left( \lambda \mI_\calH + \sum_{\tau \leq t} \alpha(\vx_\tau; \vtheta_1, \vtheta_2)\vphi(\vx_\tau)\vphi^\top(\vx_\tau) \right) \left( \vtheta_2 - \vtheta_1\right) \\
         &= G_t(\vtheta_1, \vtheta_2) \left( \vtheta_2 - \vtheta_1\right) 
    \end{align*}
    Therefore,
    \begin{align}
        \norm{ g_t(\vtheta_2) - g_t(\vtheta_1) }_{G_t^{-1}(\vtheta_1, \vtheta_2)} & = \left[g_t(\vtheta_2) - g_t(\vtheta_1) \right]^\top  \left( \vtheta_1 - \vtheta_2\right) \notag \\
        & = \left( \vtheta_2 - \vtheta_1\right) ^\top G_t \left( \vtheta_2 - \vtheta_1\right)\notag \\
        & = \norm{\vtheta_2 - \vtheta_1}_{G_t(\vtheta_1, \vtheta_2)}.\label{eq:gt_theta}
    \end{align}
    \textbf{Step 3.} Putting together the previous two steps, we can bound the $V_t$-norm over the parameters to the $V_t^{-1}$ role in the gradients,
    \begin{align*}
    \norm{\vtheta_1 -\vtheta_2}_{V_t} &  \stackrel{\text{\eqref{eq:Gt_Vt}}}{\leq}  \sqrt{\kappa}\norm{\vtheta_1 -\vtheta_2}_{G_t(\vtheta_1, \vtheta_2)}\\
         & \stackrel{\text{\eqref{eq:gt_theta}}}{\leq} \sqrt{\kappa} \norm{ g_t(\vtheta_1) - g_t(\vtheta_2) }_{G_t^{-1}(\vtheta_1, \vtheta_2)}\\
         & \stackrel{\text{\eqref{eq:Gt_Vt}}}{\leq} \kappa \norm{ g_t(\vtheta_1) - g_t(\vtheta_2) }_{V_t^{-1}}
    \end{align*}
    concluding the proof.
\end{proof}

The following two lemmas are standard results in kernelized bandits \citep[e.g.,]{srinivas2009gaussian,chowdhury2017kernelized}. We include it here for completeness. 
\begin{lemma}\label{lem:norm_conversion}
    Let $\sigma_t$ be as defined in \eqref{eq:def_sigma}. Then $\sqrt{\lambda\kappa} \norm{\vphi(\vx)}_{V_t^{-1}} = \sigma_t(\vx)$, for any $\vx \in \calX$.
\end{lemma}
\begin{proof}[Proof of \cref{lem:norm_conversion}]
    We start by stating some identities which will later be of use.  
    First note that
\[
\left( \Phi_t^\top\Phi_t + \lambda\kappa \mI_{\calH}\right) \Phi_t^\top = \Phi_t^\top  \left( \Phi_t\Phi_t^\top + \lambda\kappa \mI_{t}\right)
\]
which gives
\begin{equation}
   \Phi_t^\top \left( \Phi_t\Phi_t^\top + \lambda\kappa \mI_{t}\right)^{-1}  =  \left( \Phi_t^\top\Phi_t + \lambda\kappa \mI_{\calH}\right)^{-1}\Phi_t^\top \label{eq:conv_invs}.
\end{equation}
Moreover, by definition of $\vk_t$ we have
    \begin{align}
        &\vk_t(\vx) = \Phi_t \vphi(\vx) \label{eq:equiv_kt}
    \end{align}
    which allow us to write
\[
    \left( \Phi_t^\top\Phi_t + \lambda\kappa\mI_{\calH}\right)\vphi(\vx) = \Phi_t^\top \vk_t(\vx) + \lambda\kappa \vphi(\vx),
    \]
and obtain
\begin{align*}
\vphi(\vx) & = \left( \Phi_t^\top \Phi_t + \lambda\kappa\mI_\calH\right)^{-1} \Phi_t^{\top}\vk_t(\vx) + \lambda\kappa\left( \Phi_t^\top \Phi_t + \lambda\kappa \mI\right)^{-1} \vphi(\vx)\\
& \stackrel{\text{\eqref{eq:conv_invs}}}{=} \Phi_t^\top \left( \Phi_t\Phi_t^\top + \lambda\kappa \mI_{t}\right)^{-1} \vk_t(\vx) + \lambda\kappa\left( \Phi_t^\top \Phi_t + \lambda\kappa \mI\right)^{-1} \vphi(\vx).
\end{align*}
Given the above equation, we conclude the proof by the following chain of equations:
\begin{align*}
    k(\vx, \vx) & = \vphi^\top(\vx)\vphi(\vx)\\
    & = \left( \Phi_t^\top \left( \Phi_t\Phi_t^\top + \lambda\kappa \mI_{t}\right)^{-1} \vk_t(\vx) + \lambda\kappa\left( \Phi_t^\top \Phi_t + \lambda\kappa \mI_\calH\right)^{-1} \vphi(\vx)\right)^\top \vphi(\vx) \\
    & = \vk_t^\top(\vx) \left( \Phi_t\Phi_t^\top + \lambda\kappa \mI_{t}\right)^{-1}\Phi_t\vphi(\vx) + \lambda\kappa \vphi^\top(\vx)  \left( \Phi_t^\top \Phi_t + \lambda\kappa \mI_\calH\right)^{-1}\vphi(\vx)\\
    & \stackrel{\text{\eqref{eq:equiv_kt}}}{=}\vk_t^\top(\vx) \left(K_t + \lambda\kappa \mI_{t}\right)^{-1}\vk_t(\vx) + \lambda\kappa \vphi^\top(\vx)  V_t^{-1}\vphi(\vx)
\end{align*}
To obtain the third equation we have used the fact that for bounded operators on Hilbert spaces, the inverse of the adjoint is equal to the adjoint of the inverse \cite[e.g., Theorem 10.19,][]{axler2020measure}. Re-ordering the equation above we obtain $\sigma_t^2(\vx) = \lambda\kappa \norm{\vphi(\vx)}_{V_t^{-1}}^2$, concluding the proof.
\end{proof}
\begin{lemma}[Controlling posterior variance with information gain]\label{lem:basic_infogainsum}
For all $T\geq 1$,
\[
 \sum_{t=1}^T  \sigma^2_t(\vx_t) \leq \frac{2 \gamma_T}{\log (1+ (\lambda\kappa)^{-1})},\quad  \sum_{t=1}^T  (\sigma^\mathrm{D}_t(\vx_t))^2 \leq \frac{8\gamma^\mathrm{D}_T}{\log (1+ 4(\lambda\kappa)^{-1})}.
\]
\end{lemma}
\begin{proof}[Proof of \cref{lem:basic_infogainsum}]

By \citet[Lemma 5.3]{srinivas2009gaussian}, 
\[
\gamma_T =\max_{\vx_1, \dots \vx_T}\frac{1}{2}\sum_{t=1}^T \log (1+ (\lambda\kappa)^{-1}\sigma^2_{t-1}(\vx_t)).
\]
Following the technique in \citet[Lemma 5.4]{srinivas2009gaussian}, since $\sigma^2_t \leq 1$, then $(\lambda\kappa)^{-1}\sigma^2_t \in [0, (\lambda\kappa)^{-1}]$. 
Now for any $z \in [0, (\lambda\kappa)^{-1}]$, $z \leq C \log (1+ z)$ where $C = 1/(\lambda\kappa \log (1+ (\lambda\kappa)^{-1}))$. 
We then may write,
    \begin{align*}
        \sum_{t=1}^T  \sigma^2_t(\vx_t) & =  \sum_{t=1}^T \lambda\kappa(\lambda\kappa)^{-1}\sigma^2_t(\vx_t) \\
        & \leq  \sum_{t=1}^T \lambda\kappa C \log \left(1+ (\lambda\kappa)^{-1}\sigma^2_t(\vx_t)\right)\\
        & = \sum_{t=1}^T \frac{\log (1+ (\lambda\kappa)^{-1}\sigma^2_t(\vx_t))}{\log \left(1+ (\lambda\kappa)^{-1}\right)} 
    \end{align*}
Putting both together proves the first inequality of the lemma.
As for the dueling case, we can easily check that $\sigma_t^\mathrm{D} \leq 2$, and a similar argument yields the second inequality.
\end{proof}

%% file: appendices/2_dueling_proofs.tex
\section{Proofs for Bandits with Preference Feedback}
This section presents the proof of main results in \cref{sec:kernelized_bandits_preference_feedback}, and our additional contributions in the kernelized Preference-based setting.

\subsection{Equivalence of Preference-based and Logistic Losses}\label{app:equivalence}
We start by establishing the equivalence between the logistic loss \eqref{eq:logistic_loss} and dueling loss \eqref{eq:preference_loss}.
\begin{proof}[Proof of \cref{prop:rkhs_equival}]
By Mercer's theorem, we know that the kernel function $k$ has eigenvalue eigenfunction pairs $(\sqrt{\lambda_i}, \tilde \phi_i)$ for $i\geq 1$ where $\tilde \vphi_i$ are orthonormal.
Then $k(\vx, \vx') = \sum_{i\geq 1}\phi_i(\vx)\phi_i(\vx')$ with $\phi_i(\vx) =\sqrt{\lambda_i}\tilde \phi_i(\vx)$.
    Now applying the definition of $\kduel$, it holds that $\kduel(\vz, \vz') = \sum_{i\geq 1}\psi_i^\top(\vz)\psi_i(\vz')$ where $\psi_i(\vz) = \sqrt{\lambda_i}(\phi_i(\vx) - \phi_i(\vx'))$. It is straighforward to check that $\psi_i$ are the eigenfunctions of $\kduel$, however, they may not be orthonormal. We have,
    \begin{align*}
        \langle \psi_i , \psi_i \rangle_{L_2} & = 2\lambda_i (1-b_i^2) \\
        \langle \psi_i , \psi_j \rangle_{L_2} & = -2\sqrt{\lambda_i \lambda_j}b_ib_j
    \end{align*}
    where $b_i = \int \tilde \phi_i(\vx)\mathrm{d}(\vx)$. By the assumption of the proposition, we have $b_i=0$. However, this assumption holds automatically for all kernels commonly used in applications, e.g. any translation invariant kernel, over many domains,  since $\tilde \vphi_i$ for such kernels are a sine basis.
    
Now since $f \in \calH_k$, it may be decomposed $f = \sum_{i\geq 1} \beta_i \phi_i$ and $\norm{f}_k^2 = \sum_{i\geq 1} \beta_i^2 \leq \infty$.
And set the difference function to $h(\vx, \vx') = \sum_{i\geq 1} \beta_i \psi_i(\vz)$. We can then bound the RKHS norm of $h$ w.r.t. the kernel $\kduel$ as follows
\begin{align*}
    \norm{h}^2_{\kduel} & = \sum_{i\geq 1}\left(\frac{\langle h, \psi_i \rangle_{L_2}}{\langle \psi_i , \psi_i \rangle_{L_2}}\right)^2\\
    & =  \sum_{i\geq 1}\left(\frac{\sum_{j\geq 1}\beta_j\langle \psi_j, \psi_i \rangle_{L_2}}{2\lambda_i(1-b_i)}\right)^2\\
    & = \sum_{i\geq 1}\left(\beta_i- \frac{ b_i}{\sqrt{\lambda_i}(1-b_i)}\sum_{j\neq i}\beta_jb_j \sqrt{\lambda_j}\right)^2\\
    &\stackrel{b_i = 0}{=}\norm{f}^2_{k} \leq B^2.
\end{align*} 
Now by Mercer's theorem, $h\in \calH_{\kduel}$ since it is decomposable as a sum of $\kduel$ eigenfunctions, and attains a $B$-bounded $\kduel$-norm which we showed to be equal to $\norm{f}_k$. The other direction of the statement is proved the same way.
\end{proof}
\begin{proof}[Proof of \cref{cor:dueling_CI}]
Consider the utility function $f$ and define $h(\vx, \vx') \coloneqq f(\vx) - f(\vx')$.
Then by \cref{prop:rkhs_equival}, $h$ is in RKHS of $\kduel$ with a $\kduel$-norm bounded by $B$.
We may estimate $h$ by minimizing $\calL_{\kduel}^\mathrm{L}(\cdot; H_t)$.
Now invoking \cref{thm:func_CI} with the dueling kernel we have,
\[
 \sP\left( \forall t\geq 1, \vx, \vx' \in \calX:\, \abs{\sigmoid\left(h_t( \vx,\vx') \right)- \sigmoid\left(h( \vx,\vx')  \right)}\leq \beta^{\mathrm{D}}_t(\delta)\sigma^D_t(\vx, \vx')\right)\geq1-\delta
\] 
concluding the proof by definition of $h$.
\end{proof}

\subsection{Proof of the Preference-based Regret Bound} \label{app:dueling_regret_bound}
Recall \cref{cor:dueling_CI}, which states
\[
\abs{\sigmoid(f(\vx^\star) - f(\vx_t)) - \sigmoid(h_t(\vx^\star, \vx_t))} \leq \beta^{\mathrm{D}}_t(\delta)\sigma^D_t(\vx, \vx')
\]
with high probability simultaneously for all $(\vx, \vx')$ and $t\geq 1$.
For simplicity in notation in the rest of this section, we define $\omega_t(\vx, \vx') \coloneqq \beta^{\mathrm{D}}_t(\delta)\sigma^D_t(\vx, \vx')$ and 
\begin{align*}
    \mathrm{LCB}_t(\vx, \vx')  & = \sigmoid(h_t(\vx, \vx')) - \omega_t(\vx, \vx'),\\
    \mathrm{UCB}_t(\vx, \vx')  & =  \sigmoid(h_t(\vx, \vx')) + \omega_t(\vx, \vx').
\end{align*}
Note that $\omega_t(\vx, \vx') = \omega_t(\vx', \vx)$ by the symmetry of the dueling kernel $k^\mathrm{D}$.
Furthermore, recall the notation $h(\vx, \vx') = f(\vx) - f(\vx')$.

Before providing the proof of \cref{thm:regret_maxmin}, we derive two important inequalities. Note that $\vx_t, \vx_t' \in \calM_t$ implies that
\begin{align}
    \mathrm{UCB}_t (\vx_t, \vx_t') &\geq 0.5 \notag \\
    s(h_t (\vx_t, \vx_t')) &\geq 0.5 - \omega_t (\vx_t, \vx_t') \label{eq:mean_estimate_lowerbound}
\end{align}
Additionally note that $\mathrm{LCB}_t(\vx_t, \vx_t) = 0.5$ for all $\vx_t$, therefore, by the definition of $\vx_t'$, $\mathrm{LCB}_t(\vx_t, \vx_t') \leq 0.5$. It implies that
\begin{align}
    \mathrm{LCB}_t(\vx_t, \vx_t') &\leq 0.5 \notag \\
    s(h_t (\vx_t, \vx_t')) &\leq 0.5 + \omega_t (\vx_t, \vx_t'). \label{eq:mean_estimate_upperbound}
\end{align}
From \cref{eq:mean_estimate_lowerbound} and \cref{eq:mean_estimate_upperbound}, it follows that
\begin{equation}
    |s(h_t (\vx_t, \vx_t')) - 0.5| \leq \omega_t (\vx_t, \vx_t')  \label{eq:posterior_mean_tie_distance}
\end{equation}
furthermore,
\begin{align}
    \mathrm{UCB}_t (\vx_t, \vx_t') - 0.5 &= s(h_t (\vx_t, \vx_t')) - 0.5 + \omega_t (\vx_t, \vx_t') \notag \\
    &\leq \vert s(h_t (\vx_t, \vx_t')) - 0.5 \vert + \omega_t (\vx_t, \vx_t') \notag \\
    &\leq 2 \omega_t (\vx_t, \vx_t') \label{eq:UCB_distance_upper_bound}
\end{align}
and similarly
\begin{equation}
    0.5 - \mathrm{LCB}_t (\vx_t, \vx_t') \leq 2 \omega_t (\vx_t, \vx_t'). \label{eq:LCB_distance_upper_bound}
\end{equation}

From \cref{eq:UCB_distance_upper_bound} and \cref{eq:LCB_distance_upper_bound}, it follows that
\begin{align}
    |s(f(\vx_t) - f(\vx_t')) - 0.5| &\leq \max \{ \mathrm{UCB}_t (\vx_t, \vx_t') - 0.5, 0.5 - \mathrm{LCB}_t(\vx_t, \vx_t') \} && \text{\cref{cor:dueling_CI}} \notag \\
    &\leq 2 \omega_t (\vx_t, \vx_t'). \label{eq:mean_estimate_error_upperbound}
\end{align}
We are going to refer to \cref{eq:posterior_mean_tie_distance} and \cref{eq:mean_estimate_error_upperbound} in the following proof.

\begin{proof}[Proof of \cref{thm:regret_maxmin}]

\textbf{Step 1:}
First, we connect the term of $\vx_t'$ in the dueling regret defined in \cref{eq:dueling_regret} to that of $\vx_t$.
Note that both $s(f(\vx^*) - f(\vx_t'))$ and $s(f(\vx^*) - f(\vx_t))$ are greater than $0.5$ due to the optimality of $\vx^\star$ and the sigmoid function $s$ is concave on the interval $[0.5, \infty)$. Using the definition of concavity, we get
\begin{align}
    s(f(\vx^*) - f(\vx_t')) &\leq s(f(\vx^*) - f(\vx_t)) + \dot s(f(\vx^*) - f(\vx_t)) (f(\vx_t) - f(\vx_t')) \notag \\
    &= s(f(\vx^*) - f(\vx_t)) + s(f(\vx^*) - f(\vx_t)) s(f(\vx_t) - f(\vx^*)) (f(\vx_t) - f(\vx_t')) \notag \\
    &\leq \left( 1 + \frac{h(\vx_t, \vx_t')}{2} \right)s(f(\vx^*) - f(\vx_t)) \label{eq:second_arm_regret_ub}
\end{align}
where the second line comes from the derivative of the sigmoid function, $\dot s(x) = s(x)(1-s(x)) = s(x)s(-x)$, and in the last line we use $s(f(\vx_t) - f(\vx^*)) \leq 0.5$.

Using \cref{eq:second_arm_regret_ub}, we can upper bound the dueling regret in \cref{eq:dueling_regret} as
\begin{align}
    2r_t^D &= \sigmoid(f(\vx^\star) - f(\vx_t)) + \sigmoid(f(\vx^\star) - f(\vx'_t)) - 1 \notag \\
    &\leq \sigmoid(f(\vx^\star) - f(\vx_t)) + \left( 1+\frac{h(\vx_t, \vx_t')}{2} \right) \sigmoid(f(\vx^\star) - f(\vx_t)) - 1 \notag \\
    &\leq 2\sigmoid(f(\vx^\star) - f(\vx_t)) - 1 + \frac{h(\vx_t, \vx_t')}{2} \sigmoid(f(\vx^\star) - f(\vx_t)) \label{eq:regret_ub1}
\end{align}

\textbf{Step 2}: Next, we show that the single-step regret is bounded by $\omega_t (\vx_t, \vx_t')$.

First, consider the term $\sigmoid(f(\vx^\star) - f(\vx_t))$
\begin{align}
    \sigmoid(f(\vx^\star) - f(\vx_t)) - 0.5 &= 0.5 - \sigmoid(f(\vx_t) - f(\vx^\star)) && \text{Sigmoid definition} \notag \\
    &\leq 0.5 - \mathrm{LCB}_t(\vx_t, \vx^\star) && \text{\cref{cor:dueling_CI}} \notag \\
    &\leq 0.5 - \mathrm{LCB}_t(\vx_t, \vx'_t) && \text{Def. of $\vx'_t$} \notag \\
    &\leq \vert 0.5 - \sigmoid(h_t(\vx_t, \vx'_t)) \vert + \omega_t(\vx_t, \vx'_t) && \text{Def. of $\mathrm{LCB}_t$} \notag \\
    &\leq 2\omega_t(\vx_t, \vx'_t) && \text{\cref{eq:posterior_mean_tie_distance}} \label{eq:weak_regret_variance_ub}
\end{align}

Using \cref{eq:weak_regret_variance_ub}, we can rewrite the first term in \cref{eq:regret_ub1} to get 
\begin{equation}
    2r_t^D \leq 4\omega_t(\vx_t, \vx_t') + \frac{h(\vx_t, \vx_t')}{2} \sigmoid(f(\vx^\star) - f(\vx_t)). \label{eq:maxminlcb_regret_bound_2}
\end{equation}

It remains to bound the second term of \cref{eq:maxminlcb_regret_bound_2}. By the Mean-Value Theorem, $\exists z \in [0, h(\vx_t, \vx_t')]$ such that
\begin{align}
    \dot\sigmoid (z) (h(\vx_t, \vx_t') - 0) &= \sigmoid(h(\vx_t, \vx_t')) - f(0) \notag 
\end{align}
Now since $\kappa = \sup_{z\leq B} 1/\dot{s}(z)$ then,
\begin{align}
     h(\vx_t, \vx_t') \leq \kappa (\sigmoid(h(\vx_t, \vx_t')) - 0.5) \label{eq:delta_bound}
\end{align}

Combining \cref{eq:mean_estimate_error_upperbound} with \cref{eq:delta_bound}, it follows
\begin{equation}
    h(\vx_t, \vx_t') \leq 2\kappa\omega_t(\vx_t, \vx_t') \label{eq:utility_difference_upperbound}
\end{equation}
Using \cref{eq:utility_difference_upperbound} in \cref{eq:maxminlcb_regret_bound_2} and the fact that $\sigmoid(f(\vx^*) - f(\vx_t)) \leq 1$, we get
\begin{equation*}
    2r_t^D \leq (4+\kappa)\omega_t(\vx_t, \vx_t').
\end{equation*}
Therefore, for the cumulative dueling regret it holds 
    \begin{align*}
        R^\mathrm{D}(T)  = \sum_{t=1}^T r^\mathrm{D}_t &\leq \sqrt{T\sum_{t=1}^T (r^\mathrm{D}_t)^2} \\
        & \leq (2+\kappa/2)\beta^\mathrm{D}_T(\delta) \sqrt{T \sum_{t=1}^T (\sigma^\mathrm{D}_t)^2(\vx_t, \sqrt{\lambda\kappa})} \tag*{\text{$\beta_t(\delta) \leq \beta^\mathrm{D}_T(\delta)$}}\\
        & \leq C_3 \beta_T^\mathrm{D}(\delta)\sqrt{T\gamma^\mathrm{D}_t} \tag*{\text{Lem.~\ref{lem:basic_infogainsum}}}
    \end{align*}
    with probability greater than $1-\delta$ for all $T\geq 1$.

\end{proof}

\subsection{Extending Algorithms for Linear Dueling Bandits to Kernelized Setting} \label{appendix:algorithm_extensions_to_kernelized}
\textbf{Maximum Informative Pair Algorithm.} Proposed in \citet{saha2021optimal} for linear utilities, the \MaxInfo algorithm
similarly maintains a set of plausible maximizer arms, and picks the pair of actions that have the largest joint uncertainty, and therefore are expected to be informative. \cref{alg:max_info_kernelized} present the kernelized variant of this algorithm.
Using \cref{cor:dueling_CI}, we can show that the kernelized \MaxInfo also satisfies a $\tilde \calO(\gamma_T\sqrt{T})$ regret.

\begin{theorem}\label{thm:regret_maxinp}
        Let $\delta \in (0, 1]$ and choose the exploration coefficient 
    \smash{$\beta^{\mathrm{D}}_t(\delta)$} as defined in \cref{cor:dueling_CI}.
    Then \MaxInfo satisfies the anytime dueling regret guarantee of 
\begin{align*}
\sP\left(\forall T\geq 0: R^{\mathrm{D}}(T) \leq C_2\beta^\mathrm{D}_T(\delta)\sqrt{T\gamma^\mathrm{D}_T}\right) \geq 1-\delta
\end{align*}
where $\gamma^\mathrm{D}_T$ is the $T$-step information gain of kernel $k^\mathrm{D}$ and $C_2 = 4/\sqrt{\log (1+ 4(\lambda\kappa)^{-1})}$.
\end{theorem}
\begin{proof}[Proof of \cref{thm:regret_maxinp}]
When selecting $(\vx_t, \vx'_t)$ according to \cref{alg:max_info_kernelized}, we choose the pair via
\begin{equation}\label{eq:acq_rule}
   \vx_t, \vx'_t = \argmax_{\vx, \vx' \in \calM_t} \omega_t(\vx, \vx') 
\end{equation}
where action space is restricted to $\calM_t$ and therefore,
\begin{equation} \label{eq:plaus_max_cond}
    \begin{split}
        \sigmoid(h_t(\vx^\star, \vx_t)) & \leq 1/2 + \omega_t(\vx_t, \vx^\star)\\
        \sigmoid(h_t(\vx^\star, \vx'_t)) & \leq 1/2 + \omega_t(\vx'_t, \vx^\star)\\
    \end{split}
\end{equation}
where we have used the identity $\sigmoid(-z) = 1-\sigmoid(z)$.
Simultaneously for all $t\geq 1$, we can bound the single-step dueling regret with probability greater than $1-\delta$
\begin{align*}
    2r^\mathrm{D}_t & = \sigmoid(f(\vx^\star) - f(\vx_t)) + \sigmoid(f(\vx^\star) - f(\vx'_t)) -1\\
& \leq \sigmoid(h_t(\vx^\star, \vx_t)) + \omega_t(\vx^\star, \vx_t) + \sigmoid(h_t(\vx^\star, \vx'_t)) + \omega_t(\vx^\star, \vx'_t) -1 \tag{w.h.p.}\\
& \leq 2\left(\omega_t(\vx^\star, \vx_t) + \omega_t(\vx^\star, \vx'_t)\right)\tag*{\text{Eq.~\eqref{eq:plaus_max_cond}}}\\
& \leq 4 \omega_t(\vx_t, \vx'_t))\tag*{\text{Eq.~\eqref{eq:acq_rule}}}
    \end{align*}
where for the first inequality we have invoked \cref{cor:dueling_CI}.
Then for the regret satisfies
    \begin{align*}
        R^\mathrm{D}(T)  = \sum_{t=1}^T r^\mathrm{D}_t &\leq \sqrt{T\sum_{t=1}^T (r^\mathrm{D}_t)^2} \\
        & \leq 2\beta^\mathrm{D}_T(\delta) \sqrt{T \sum_{t=1}^T (\sigma^\mathrm{D}_t)^2(\vx_t, \sqrt{\lambda\kappa})} \tag*{\text{$\beta_t(\delta) \leq \beta^\mathrm{D}_T(\delta)$}}\\
        & \leq C_2 \beta_T^\mathrm{D}(\delta)\sqrt{T\gamma^\mathrm{D}_t} \tag*{\text{Lem.~\ref{lem:basic_infogainsum}}}
    \end{align*}
    with probability greater than $1-\delta$ for all $T\geq 1$.
\end{proof}

\begin{algorithm}[t]
\caption{\MaxInfo - Kernelized Variant \label{alg:max_info_kernelized}}
\begin{algorithmic}
\State \textbf{Input} $(\beta^\mathrm{D}_t)_{t\geq 1}$. %
\For{$t \geq 1$}
\State Play the most informative pair via
\[
\vx_t, \vx'_t = \arg\max_{\vx, \vx'\calM_t} \sigma^\mathrm{D}_t(\vx, \vx')
\]
\State Observe $y_t$ and append history.
\State Update $h_{t+1}$ and $\sigma^\mathrm{D}_{t+1}$ and the set of plausible maximizers
\[\calM_{t+1} = \{\vx \in \calX\vert\, \forall \vx' \in \calX:\,  \sigmoid(h_{t+1}(\vx,\vx')) + \beta^\mathrm{D}_{t+1}\sigma^\mathrm{D}_{t+1}(\vx, \vx') > 1/2  \}.\] 
\EndFor
\end{algorithmic}
\end{algorithm}
\vspace{-10pt}

\textbf{Dueling Information Directed Sampling (\IDS) Algorithm.} To choose actions at each iteration $t$, \MaxInfo and \dalgo require solving an optimization problem on $\calX \times \calX$.
The Dueling \IDS approach addresses this issue and presents an algorithm which requires solving an optimization problem on $\calX \times [0,1]$ and is computationally more efficient when $d_0>1$. 
This work considers kernelized utilities, however, assumes the probability of preference itself is in an RKHS and solves a kernelized ridge regression problem to estimate the probability $s(h(\vx, \vx')$.
In the following, we present an improved version of this algorithm, by considering the preference-based loss \eqref{eq:preference_loss} for estimating the utility function. We modify the algorithm and the theoretical analysis to accommodate this.\looseness -1

Consider the sub-optimality gap $\Delta(\vx) \coloneqq h(\vx^\star,\vx)$ for an action $\vx \in \calX$. We may estimate this gap using the reward estimate maximizer $\hat \vx_t^\star \coloneqq \argmax_{\vx \in \calX}f_t(\vx)$. Suppose we choose $\hat\vx^\star_t$ as one of the actions. Then $u_t$ shows an optimistic estimate of the highest obtainable reward at this step:
\[
u_t \coloneqq \max_{\vx \in \calX} h(\vx, \hat \vx_t^\star) + \tilde \beta_t\sigma_t^D(\vx, \vx_t^\star).
\]
where $\tilde \beta_t$ is the exploration coefficient. We bound $\Delta(\vx)$ by the estimated gap 
\begin{equation}
    \label{eq:def_gap}
    \hat \Delta_t(\vx) \coloneqq u_t + h_t(\hat \vx^\star_t, \vx)
\end{equation}
and show its uniform validity in 
\cref{lem:gap_bound}.
We can now propose the Kernelized Logistic IDS algorithm with preference feedback in \cref{alg:kernel_ids}, as a variant of the algorithm of \citeauthor{kirschner2021bias}.\looseness-1
\begin{theorem}\label{thm:ids_regret}
            Let $\delta \in (0, 1]$ and for all $t\geq 1$, set the exploration coefficient as $\tilde \beta_t =\beta_t^\mathrm{D}(\delta)/L$.
    Then \cref{alg:kernel_ids} satisfies the anytime cumulative dueling regret guarantee of 
\begin{align*}
\sP\left(\forall T\geq 0: R^{\mathrm{D}}(T) = \calO\left(\beta_T^\mathrm{D}(\delta)\sqrt{T(\gamma_T + \log 1/\delta)}\right) \right) \geq 1-\delta.
\end{align*}
\end{theorem}

\begin{algorithm}[t]
\caption{Preference-based IDS - Kernelized Logistic Variant \label{alg:kernel_ids} }
\begin{algorithmic}
\State \textbf{Initialize} Set $(\beta_t)_{t\geq 1}$ according to \cref{thm:func_CI}.
\For{$t \geq 1$}
\State Find a greedy action via fixing any point $x_\mathrm{null} \in \calX$ and maximizing
\[
  \vx_t^{(1)} = \hat \vx^\star_t = \argmax_{x \in \calX} h_t(x, x_{\mathrm{null}}). 
\]
\State Update $u_t$ and $\hat \Delta_t(\vx)$ acc. to \eqref{eq:def_gap}.
\State Find an informative action and the probability of selection via
\[
\vx^{(2)}_t, p_t = \argmin_{\substack{\vx \in \calX\\p \in [0,1]}} \frac{\Big((1-p)u_t + p \hat \Delta_t(\vx)\Big)^2}{p\log\left(1+(\lambda \kappa)^{-1} \big(\sigma_t^D(\vx^{(1)}_t, \vx)\big)^2\right)}.
\]
\State Draw $\alpha_t \sim \mathrm{Bern}(p_t)$.
\State \textbf{if} $\alpha_t = 1$ \textbf{then} choose pair $(\vx_t, \vx_t') = (\vx^{(1)}_t, \vx^{(2)}_t) $ \textbf{else} choose  $(\vx_t, \vx_t') = (\vx^{(1)}_t, \vx^{(1)}_t)$.
\State Observe $y_t$ and append history.
\State Update $h_{t+1}$ and $\sigma^\mathrm{D}_{t+1}$.
\EndFor
\end{algorithmic}
\end{algorithm}
\vspace{-10pt}

%% file: appendices/3_ids_proof.tex
 \begin{proof}[Proof of \cref{thm:ids_regret}]
Our approach closely follows the proof of \citet[Theorem 1]{kirschner2021bias}.
Let $\calP(\cdot)$ show the set of continuous probability distributions over a domain.
    Define the expected average gap for a policy $\mu \in \calP(\calX \times \calX)$
    \[
    \hat \Delta_t(\mu) \coloneqq \frac{1}{2}\Exp_{\vx, \vx' \sim \mu} \hat \Delta_t(\vx) + \hat \Delta_t(\vx')
    \]
    and the expected information ratio as
    \[
    \Xi_t(\mu) \coloneqq \frac{\hat \Delta^2_t(\mu)}{\Exp_{\vx, \vx' \sim \mu}\log\left( 1+ (\lambda \kappa)^{-1}\big(\sigma^D_t(\vx, \vx')\big)^2\right)}.
    \]
    \cref{alg:kernel_ids} draws actions via $\mu_t= (1-p_t)\delta_{(\vx^{(1)}_t, \vx^{(1)}_t)}+ p_t \delta_{(\vx^{(1)}_t, \vx^{(2)}_t)}$, where $\delta_{(\vx, \vx')}$ denotes the Dirac delta.
    Then by \citet[Lemma 1]{kirschner2020information}, 
    \[
    \frac{1}{2}\sum_{t=1}^T h(\vx^\star, \vx_t) + h(\vx^\star, \vx'_t)
    \leq \sqrt{\sum_{t=1}^T\Xi_t(\mu_t) \left(\gamma_T + \calO(\log 1/\delta)\right)} + \calO(\log T/\delta)
    \]
    which allows us to bound the regret with probability greater than $1-\delta$ as 
    \begin{equation}
           \label{eq:generic_ids_regret}
        R^\mathrm{D}(T) \leq L\sqrt{\sum_{t=1}^T\Xi_t(\mu_t)\left(\gamma_T + \calO(\log 1/\delta)\right)} + \calO(L\log T/\delta)
    \end{equation}
    since $\sigmoid(\cdot)$ with its domain restricted to $[-2B, 2B]$ is $L$-Lipschitz.
    It remains to bound $\Xi_t(\mu_t)$, the expected information ratio for \cref{alg:kernel_ids}.
    Now by definition of $\mu_t$
    \begin{align*}
            2\hat\Delta_t(\mu_t) & = (2-p_t)\hat \Delta_t(\vx^{(1)}_t) + p_t\Delta_t(\vx^{(2)}_t)\\
            & = (2-p_t)\left(u_t + h_t(\vx_t^{(1)}, \vx_t^{(1)}) \right)+ p_t\Delta_t(\vx^{(2)}_t)\\
            & = 2(1-p_t)u_t + p_t(\hat \Delta_t(\vx_t^{(2)}) + u_t),
    \end{align*}
    and similarly
    \begin{align*}
        \Exp_{ \mu_t}\log\left( 1+ \tfrac{\sigma^D_t(\vx, \vx')^2}{\lambda \kappa}\right)& = (1-p_t)\log\left( 1+ \tfrac{\sigma^D_t(\vx_t^{(1)},\vx_t^{(1)})^2}{\lambda \kappa}\right) + p_t\log\left( 1+ \tfrac{\sigma^D_t(\vx_t^{(1)},\vx_t^{(2)})^2}{\lambda \kappa}\right)\\
        & = p_t\log\left( 1+ (\lambda \kappa)^{-1}\sigma^D_t(\vx_t^{(1)},\vx_t^{(2)})^2\right)\tag{\text{$\sigma_t^D(\vx, \vx) = 0$}}
    \end{align*}
allowing us to re-write the expected information ratio as
    \begin{align}
        \Xi_t(\mu_t) & = \frac{\left( 2(1-p_t)u_t + p_t(\hat \Delta_t(\vx_t^{(2)}) + u_t)\right)^2}{4p_t\log\left( 1+ (\lambda \kappa)^{-1}\sigma^D_t(\vx_t^{(1)},\vx_t^{(2)})^2\right)}\notag\\
        & \leq \frac{\left( (1-p_t)u_t + p_t\hat \Delta_t(\vx_t^{(2)})\right)^2}{p_t\log\left( 1+ (\lambda \kappa)^{-1}\sigma^D_t(\vx_t^{(1)},\vx_t^{(2)})^2\right)}\tag{\text{$u_t \leq \hat \Delta_t(\vx)$}}\\
        & = \min_{\vx, p}\frac{\left( (1-p)u_t + p\hat \Delta_t(\vx)\right)^2}{p\log\left( 1+ (\lambda \kappa)^{-1}\sigma^D_t(\vx_t^{(1)},\vx)^2\right)}\tag*{\text{Def.~($p_t, \vx_t^{(2)}$)}}\\
        & \leq \min_{\vx}\frac{\hat \Delta^2_t(\vx)}{\log\left( 1+ (\lambda \kappa)^{-1}\sigma^D_t(\vx_t^{(1)},\vx)^2\right)}.\tag*{\text{Set $p=1$}}
    \end{align}
Now consider the definition of $u_t$ and let $\vz_t$ denote the action for which $u_t$ is achieved, i.e. $\vz_t = \arg\max h(\vx, \hat\vx_t^\star) + \bar \beta_t(\delta)\sigma_t^\mathrm{D}(\vx, \hat \vx_t^\star)$. Then
\[
\hat \Delta_t(\vz_t) = h(\hat\vx_t^\star, \vz_t) +  \bar \beta_t(\delta)\sigma_t^\mathrm{D}(\vz_t, \hat \vx_t^\star) + h(\vz_t, \hat\vx_t^\star) = \bar \beta_t(\delta)\sigma_t^\mathrm{D}(\vx, \hat \vx_t^\star), 
\]
therefore using the above chain of equations we may write
\begin{align}
    \Xi_t(\mu_t) & \leq \min_{\vx}\frac{\hat \Delta^2_t(\vx)}{\log\left( 1+ \sigma^D_t(\vx_t^{(1)},\vx)^2\right)} \notag\\
    & \leq \frac{\hat \Delta^2_t(\vz_t)}{\log\left( 1+ (\lambda \kappa)^{-1}\sigma^D_t(\vx_t^{(1)},\vz_t)^2\right)}\notag\\
    & \leq \frac{ \bar \beta^2_t(\delta)\sigma_t^\mathrm{D}(\vz_t \hat \vx_t^\star)^2}{\log\left( 1+ (\lambda \kappa)^{-1}\sigma^D_t(\vx_t^{(1)},\vz_t)^2\right)}\notag\\
    & \leq \frac{4\bar \beta^2_t(\delta)}{\log\left(1+ 4(\lambda \kappa)^{-1}\right)} \label{eq:inf_ratio_bound}
\end{align}
where last inequality holds due to the following argument.
Recall that $k(\vx, \vx) \leq 1$, implying that $\sigma_t^\mathrm{D}(\vx, \vx')^2 \leq 4$ and therefore $\log(1+\sigma_t^\mathrm{D}(\vx, \vx')^2)\geq \log(1+ (\lambda\kappa)^{-1})\sigma_t^\mathrm{D}(\vx, \vx')^2/4$, similar to \cref{lem:basic_infogainsum}.
To conclude the proof, from \eqref{eq:generic_ids_regret} and \eqref{eq:inf_ratio_bound} it holds that
\begin{align*}
     R^\mathrm{D}(T)& \leq L\sqrt{\sum_{t=1}^T\Xi_t(\mu_t)\left(\gamma_T + \calO(\log 1/\delta)\right)} + \calO(L\log T/\delta)\\
     & \leq L\sqrt{\sum_{t=1}^T\frac{4\bar \beta^2_t(\delta)}{\log\left(1+ 4(\lambda \kappa)^{-1}\right)}\left(\gamma_T + \calO(\log 1/\delta)\right)}+ \calO(L\log T/\delta)\\
     & \leq L\sqrt{\frac{4T\bar \beta^2_T(\delta)}{\log\left(1+ 4(\lambda \kappa)^{-1}\right)}\left(\gamma_T + \calO(\log 1/\delta)\right)}+ \calO(L\log T/\delta)\\
     &  = \calO\left(\beta_T^\mathrm{D}(\delta)\sqrt{T(\gamma_T + \log 1/\delta)}\right)
\end{align*}
with probability greater than $1-\delta$, simultaneously for all $T\geq 1$.
\end{proof}
\subsection{Helper Lemmas for \texorpdfstring{\cref{appendix:algorithm_extensions_to_kernelized}}{}}
\begin{lemma}\label{lem:gap_bound}
    Let $0<\delta <1$ and $f \in \calH_k$. Suppose $\sup_{a \leq B}\dot{\sigmoid}(a) = L$ and $\sup_{a \leq B} 1/\dot{\sigmoid}(a) = \kappa$.
Then
    \[
    \sP(\forall t\geq 0, \vx\in \calX:\, \Delta(\vx) \leq 2 \hat\Delta_t(\vx)) \geq 1-\delta.
    \]
\end{lemma}
\begin{proof}[Proof of \cref{lem:gap_bound}]
Note that for any three inputs $\vx_1, \vx_2, \vx_3$
\begin{equation}
    \label{eq:h_property}
    h(\vx_1, \vx_3) = h(\vx_1, \vx_2) + h(\vx_2, \vx_3).
\end{equation}
Therefore, from the definition of the estimated gap get
\begin{align}
   \label{eq:basic_Delta_bound}
    \hat \Delta_t(\vx) & = \max_{\vz \in \calX} h(\vz, \hat \vx_t^\star)+ h_t(\hat \vx^\star_t, \vx) + \bar \beta_t(\delta) \sigma_t^D(\vz, \hat \vx_t^\star) \notag\\
    & = \max_{\vz \in \calX} h(\vz, \vx)+ \bar \beta_t(\delta) \sigma_t^D(\vz, \hat \vx_t^\star)\notag\\
    & \geq h(\vx, \vx) + \bar \beta_t(\delta) \sigma_t^D(\vx, \vx_t^\star)\notag\\
    & = \bar \beta_t(\delta) \sigma_t^D(\vx, \vx_t^\star).
\end{align}
Then going back to the definition of the true gap we may write
    \begin{align*}
        \Delta(\vx) & = \max_{\vz \in \calX} h(\vz,\vx)\\
        & = \max_{\vz \in \calX} h(\vz,\hat \vx_t^\star) + h(\hat \vx_t^\star,\vx)\tag*{\text{Eq.~\eqref{eq:h_property}}}\\
        & \stackrel{\mathrm{w.h.p.}}{\leq}   \max_{\vz \in \calX} h^P_t(\vz,\hat \vx_t^\star) + h_t(\hat \vx_t^\star,\vx) + \bar\beta_t(\delta)\Big(\sigma_t^D(\vz, \hat\vx_t^\star) + \sigma_t^D( \hat \vx_t^\star, \vx)\Big) \tag*{ Lem.~\ref{lem:aux_CI}}\\
        & = u_t + h^P_t(\hat \vx_t^\star,\vx) +  \bar\beta_t(\delta)\sigma_t^D( \hat \vx_t^\star, \vx)\tag*{\text{Def.~$u_t$}}\\
        & = \hat \Delta_t(\vx) +  \bar\beta_t(\delta)\sigma_t^D( \vx_t^\star, \vx) \tag*{\text{Def.~$\hat \Delta_t(\vx)$}}\\
        & \leq 2 \hat \Delta_t(\vx) \tag*{Eq.~\eqref{eq:basic_Delta_bound}}
    \end{align*}
    with probability greater than $1-\delta$.
\end{proof}

\begin{lemma}\label{lem:aux_CI}
        Assume $f\in \calH_k$. Suppose $\sup_{a \leq B} 1/\dot{\sigmoid}(a) = \kappa$.
Then for any $0<\delta <1$
\[
\sP\left( \forall t\geq1, x\in\calX:\, \abs{h(\vx,\vx') - h^P_t(\vx,\vx') } \leq \bar\beta_t(\delta)\sigma^D_t(\vx,\vx'; \sqrt{\lambda\kappa})\right) \geq 1-\delta
\]    
where
\[
\bar \beta_t(\delta) \coloneqq 2B + \sqrt{\frac{\kappa}{\lambda}}\sqrt{2\log 1/\delta + 2\gamma_t(\sqrt{\lambda\kappa})}.
\]
\end{lemma}
\begin{proof}[Proof of \cref{lem:aux_CI}] This lemma is effectively a weaker parallel of \cref{cor:dueling_CI}. We have
    \begin{align*}
        \abs{h(\vx,\vx') - h^P_t(\vx,\vx') } & = \abs{f(\vx,) - f(\vx') - (f^P_t(\vx,) - f^P_t(\vx')) }\\
        & = \abs{\vpsi^\top(\vx, \vx')(\vtheta^\star - \vtheta^P_t)}\\
     & \leq \norm{\vpsi(\vx, \vx')}_{(V^D_t)^{-1}} \norm{\vtheta^\star - \vtheta^P_t}_{V^D_t}\\
       & \stackrel{\text{w.h.p.}}{\leq}  \sqrt{\lambda\kappa} \bar \beta_t(\delta)\norm{\vpsi(\vx, \vx')}_{(V^D_t)^{-1}}\tag*{\text{Lem.~\ref{lem:CS_valid}}} \\
       & \leq \bar \beta_t(\delta)\sigma^D_t(\vx, \sqrt{\lambda\kappa})\tag*{\text{Lem.~\ref{lem:norm_conversion}}}
    \end{align*}
    where the third to last inequality holds with probability greater than $1-\delta$ , but the rest of the inequalities hold deterministically.
\end{proof}

%% file: appendices/4_additional_experiments.tex
\section{Details of Experiments}
\label{appendix:numerical_experiments}
\vspace{-10pt}
\begin{figure}[H]
    \centering
    \includegraphics[width=0.7\textwidth]{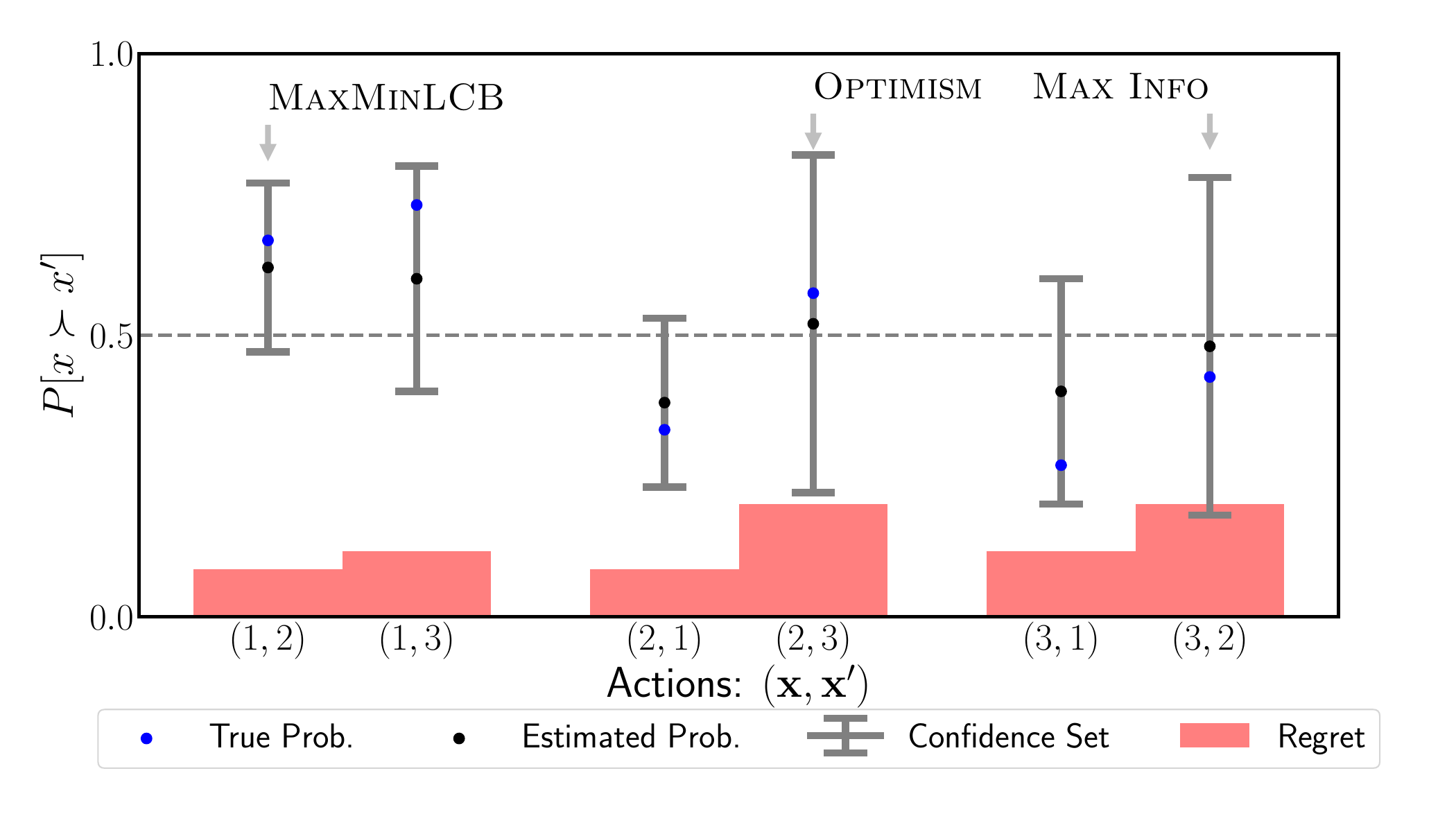}
    \vspace{-15pt}
    \caption{
    Confidence sets for an illustrative problem with $3$ arms at a single time step.
    Annotated arrows highlight the action selection for three common approaches.
    \dalgo selects the action pair $(1,2)$ with the least regret.
    Upper-bound maximization ($\textsc{Optimism}$) and information maximization ($\textsc{Max Info}$) choose sub-optimal arms.
    }
    \label{fig:action_selection}
    \vspace{-15pt}
\end{figure}

\textbf{Test Environments.}
\label{appendix:optimization_functions}
We use a wide range of target functions common to the optimization literature \citep{jamil2013literature}, to evaluate the robustness of \dalgo. The results are reported in \cref{table:regret_comparison} and \cref{table:logistic_comparison}.
Note that for the experiments we negate them all to get utilities.
We use a uniform grid of $100$ points over their specified domains and scale the utility values to $[-3, 3]$.
\begin{itemize}
    \item Ackley: $\calX = [-5, 5]^d, d=2$
    \vspace{-2pt}
    \begin{equation*}
        f(\vx) = -20\exp\left(-0.2\sqrt{\frac{1}{d}\sum_{i=1}^d x_i^2}\right) - \exp\left(\frac{1}{d}\sum_{i=1}^d \cos (2\pi x_i)\right) + 20 + \exp(1)
    \end{equation*}
    \item Branin: $\calX = [-5, 10] \times [0, 15]$
    \begin{equation*}
        f(\vx) = \left(x_2 - \frac{5.1}{4\pi^2}x_1^2 + \frac{5}{\pi}x_1 - 6\right)^2 + 10\left(1 - \frac{1}{8\pi}\right)\cos(x_1) + 10
    \end{equation*}
    \item Eggholder: $\calX = [-512, 512]^2$
    \begin{equation*}
        f(\vx) = -(x_2 + 47) \sin\left(\sqrt{|x_2 + \frac{x_1}{2} + 47|}\right) - x_1 \sin\left(\sqrt{|x_1 - (x_2 + 47)|}\right)
    \end{equation*}
    \item Hölder: $\calX = [-10, 10]^2$
    \begin{equation*}
        f(\vx) = -|\sin(x_1)\cos(x_2)\exp\left(|1 - \frac{\sqrt{x_1^2 + x_2^2}}{\pi}|\right)|
    \end{equation*}
    \item Matyas: $\calX = [-10, 10]^2$
    \begin{equation*}
        f(\vx) = 0.26(x_1^2 + x_2^2) - 0.48x_1x_2
    \end{equation*}
    \item Michalewicz: $\calX = [0, \pi]^d, d = 2, m = 10$
    \begin{equation*}
        f(\vx) = -\sum_{i=1}^d \sin(x_i) \sin^{2m} \left( \frac{ix_i^2}{\pi} \right)
    \end{equation*}
    \item Rosenbrock: $\calX = [-5, 10]^2$
    \begin{equation*}
        f(\vx) = \sum_{i=1}^{d-1} \left[ 100(x_{i+1} - x_i^2)^2 + (x_i - 1)^2 \right]
    \end{equation*}
\end{itemize}
\vspace{-10pt}

\begin{algorithm}[h!]
\caption{\Doubler \citep{ailon2014reducing} \label{alg:Doubler}}
\begin{algorithmic}
    \State \textbf{Input} $(\beta^\mathrm{D}_t)_{t\geq 1}$. %
    \State Let $\calL$ be any action from $\calX$
    \For{$t \geq 1$}
    \For{$j = 1,\dots,2^t$}
        \State Select $\vx_t'$ uniformly randomly from $\calL$
        \State Select $\vx_t = \argmax_{\vx \in \calM_t} \sigmoid(h_t(\vx,\vx_t')) + \beta^\mathrm{D}_t \sigma^\mathrm{D}_t(\vx, \vx_t')$
        \State Observe $y_t$ and append history.
        \State Update $h_{t+1}$ and $\sigma^\mathrm{D}_{t+1}$
    \EndFor
    \State $\calL \leftarrow$ the multi-set of actions played as $\vx_t'$ in the last for-loop over index $j$
    \EndFor
\end{algorithmic}
\end{algorithm}
\vspace{-10pt}
\begin{algorithm}[h!]
\caption{\MultiSBM \citep{ailon2014reducing} \label{alg:MultiSBM}}
\begin{algorithmic}
    \State \textbf{Input} $(\beta^\mathrm{D}_t)_{t\geq 1}$. %
    \For{$t \geq 1$}
    \State Set $\vx_t \gets \vx_{t-1}'$
    \State Select $\vx_t' = \argmax_{\vx \in \calM_t} \sigmoid(h_t(\vx,\vx_t)) + \beta^\mathrm{D}_t \sigma^\mathrm{D}_t(\vx, \vx_t)$
    \State Observe $y_t$ and append history.
    \State Update $h_{t+1}$ and $\sigma^\mathrm{D}_{t+1}$ and the set of plausible maximizers
    \[\calM_{t+1} = \{\vx \in \calX\vert\, \forall \vx' \in \calX:\,  \sigmoid(h_{t+1}(\vx,\vx')) + \beta^\mathrm{D}_{t+1}\sigma^\mathrm{D}_{t+1}(\vx, \vx') > 1/2  \}.\] 
    \EndFor
\end{algorithmic}
\end{algorithm}
\vspace{-10pt}
\begin{algorithm}[h!]
\caption{\RUCB \citep{zoghi2014relative} \label{alg:RUCB}}
\begin{algorithmic}
    \State \textbf{Input} $(\beta^\mathrm{D}_t)_{t\geq 1}$. %
    \For{$t \geq 1$}
    \State Select $\vx_t'$ uniformly randomly from $\calM_t$
    \State Select $\vx_t = \argmax_{\vx \in \calM_t} \sigmoid(h_t(\vx,\vx_t')) + \beta^\mathrm{D}_t \sigma^\mathrm{D}_t(\vx, \vx_t')$
    \State Observe $y_t$ and append history.
    \State Update $h_{t+1}$ and $\sigma^\mathrm{D}_{t+1}$ and the set of plausible maximizers
    \[\calM_{t+1} = \{\vx \in \calX\vert\, \forall \vx' \in \calX:\,  \sigmoid(h_{t+1}(\vx,\vx')) + \beta^\mathrm{D}_{t+1}\sigma^\mathrm{D}_{t+1}(\vx, \vx') > 1/2  \}.\] 
    \EndFor
\end{algorithmic}
\end{algorithm}
\textbf{Acquisition Function Maximization.}
In our computations, to eliminate additional noise coming from approximate solvers, we use an exhaustive search over the domain for the action selection of \algo, \dalgo, and other presented algorithms.
For the numerical experiments presented in this paper, we do not consider this as a practical limitation.
Due to our efficient implementation in JAX, this optimization step can be carried out in parallel and seamlessly support accelerator devices such as GPUs and TPUs.

\textbf{Hyper-parameters for Logistic Bandits.} We set $\delta=0.1$ for all algorithms. For \GPUCB and \algo, we set $\beta=1$, and $0.25$ for the noise variance. We use the Radial Basis Function (RBF) kernel and choose the variance and length scale parameters from $[0.1, 1.0]$ to optimize their performance separately. For \algo, we tuned $\lambda$, the $L2$ penalty coefficient in \cref{eq:logistic_nll}, on the grid $[0.0, 0.1, 1.0, 5.0]$ and $B$ on $[1.0, 2.0, 3.0]$. The hyper-parameter selections were done for each algorithm separately to create a fair comparison.\looseness-1

\textbf{Hyper-parameters for Preference-based Bandits.} We tune the same parameters of \algo for the preference feedback bandit problem on the following grid: $\lambda \in [0, 0.1, 1]$, $B \in [1, 2, 3]$, and $[0.1, 1]$ for the kernel variance and length scale. The same hyper-parameters are tuned separately for every baseline . 

\textbf{Pseudo-code for Baselines.}
\cref{alg:Doubler}, \cref{alg:MultiSBM}, and \cref{alg:RUCB} described the baselines used for the benchmark of \cref{sec:experiment_dueling}. \MaxInfo and \IDS are defined in \cref{alg:max_info_kernelized} and \cref{alg:kernel_ids}, respectively, in \cref{appendix:algorithm_extensions_to_kernelized} alongside with their theoretical analysis. We note that \Doubler includes an internal for-loop, therefore, we adjusted the time-horizon $T$ such that it observes the same number of feedback $y_t$ as the other algorithms for a fair comparison.

\textbf{Computational Resources and Costs.}
\label{appendix:experiments_computational_resources}
We ran our experiments on a shared cluster equipped with various NVIDIA GPUs and AMD  EPYC CPUs. Our default configuration for all experiments was a single GPU with 24 GB of memory, 16 CPU cores, and 16 GB of RAM. Each experiment of the $11$ configurations reported in \cref{sec:experiment_dueling} ran for about $12$ hours and the experiment reported in \cref{sec:logistic_bandit_experiment} ran for 5 hours. The total computational cost to reproduce our results is around 140 hours of the default configuration. Our total computational costs including the failed experiments are estimated to be 2-3 times more.

\subsection{Yelp Experiment}
\label{appendix:yelp_experiments}

We filter the Yelp data\footnote{Source: \href{https://www.yelp.com/dataset}{https://www.yelp.com/dataset}} for restaurants in Philadelphia, USA with at least $500$ reviews and users who reviewed at least $90$ restaurants. The final dataset includes $275$ restaurants, $20$ users, and a total of $2563$ reviews.
We define the action space $\calX$ by assigning to each restaurant their respective $32$-dimensional embedding of their reviews, i.e., $\calX \subseteq \sR^{32}$.
For each restaurant, we concatenate all reviews in the filtered dataset and we use the \textsc{text-embedding-3-large} OpenAI embedding model \footnote{Source: \href{https://platform.openai.com/docs/guides/embeddings}{https://platform.openai.com/docs/guides/embeddings}} to retrieve the embeddings.
The Yelp dataset provides utility values for users in the form of ratings on the scale of $1$ to $5$, however, not all users rated every restaurant.
We use collaborative filtering to estimate the missing reviews \citep{schafer2007collaborative}. For each user, these values are used as the utilities for restaurants during the simulation.
Experiments are conducted separately for each user, therefore, utility values are not aggregated but the action space is identical for each experiment.

Note that we do not assume any explicit functional form of the utility functions $f$ that we calibrate to this data. Instead, the actions space $\calX$ and utility values are derived separately from the dataset. Regardless, the results presented in \cref{sec:yelp} show that our kernelized method achieves good performance on this task.

\section{Additional Experiments}
\label{appendix:additional_experiments}
\begin{figure}[h]
    \centering
    \begin{subfigure}[b]{0.48\textwidth}
        \centering
        \includegraphics[width=1.0\textwidth]{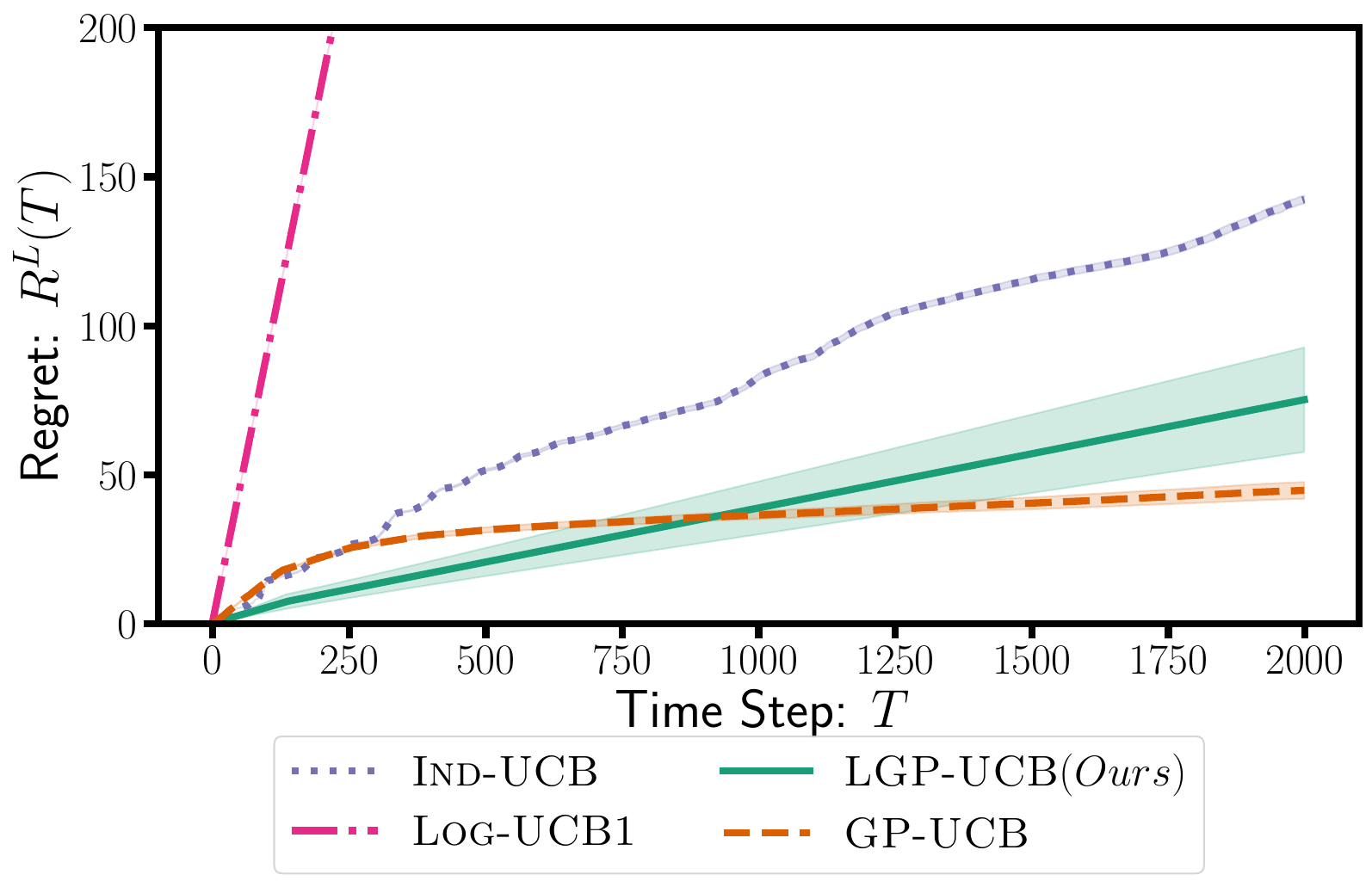}   
            
    \end{subfigure}
    \begin{subfigure}[b]{0.505\textwidth}
        \centering
        \includegraphics[width=1.0\textwidth]{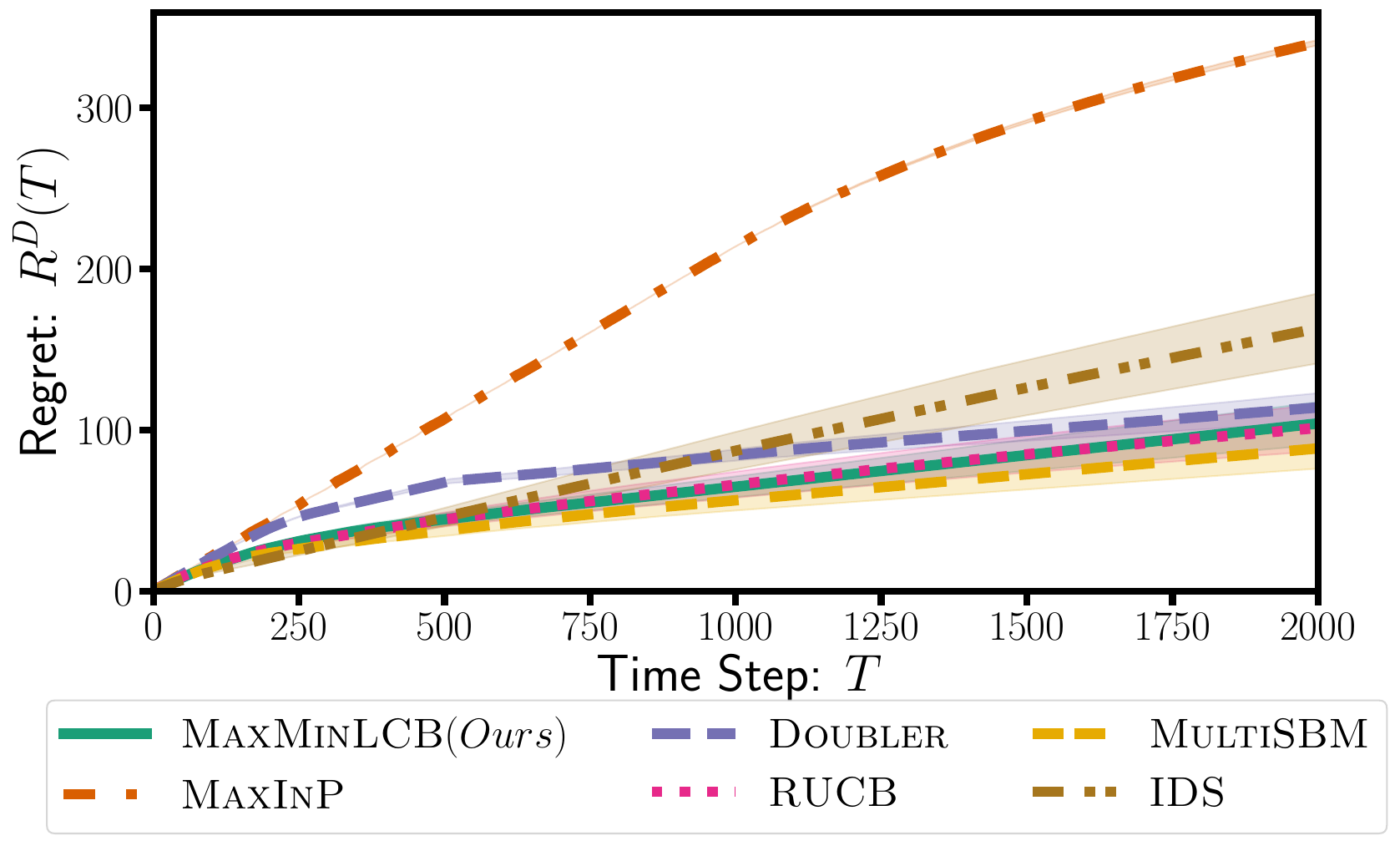}
       
    \end{subfigure}
    \caption{Regret with Branin utility function with logistic (left) and preference (right) feedback. \label{fig:branin_regret}}
    \vspace{-20pt}
\end{figure}

In this section, we provide \cref{table:logistic_comparison} that details our logistic bandit benchmark, complementing the results in \cref{sec:logistic_bandit_experiment}. 
\cref{fig:branin_regret} and \cref{fig:all} show the logistic and dueling regret of additional test functions, complementing the results of \cref{table:regret_comparison}.

\begin{table}[h]
\vspace{-10pt}
  \caption{Benchmarking $R^{\mathrm{L}}_T$ for a variety of test utility functions, $T=2000$.}
  \label{table:logistic_comparison}
  \centering
  \begin{tabular}{lcccc}
    \toprule
    $f$ & \algo  & \GPUCB & \UCB & \LOGUCB \\
    \midrule
    Ackley & $\boldsymbol{23.97} \pm 1.54$ & $96.35 \pm 1.27$ & $479.63 \pm 3.42$ & $1810.30 \pm 0.00$ \\
    Branin & $75.23 \pm 17.51$ & $\boldsymbol{44.81} \pm 2.81$ & $142.37 \pm 1.33$ & $1810.30 \pm 0.00$ \\
    Eggholder & $167.11 \pm 31.26$ & $\boldsymbol{152.34} \pm 4.28$ & $559.56 \pm 4.15$ & $1041.00 \pm 0.00$ \\
    Hoelder & $\boldsymbol{57.35} \pm 10.23$ & $150.41 \pm 9.64$ & $426.28 \pm 2.94$ & $105.64 \pm 4.88$ \\
    Matyas & $\boldsymbol{36.64} \pm 8.77$ & $50.21 \pm 2.07$ & $137.98 \pm 1.21$ & $920.48 \pm 0.57$ \\
    Michalewicz & $283.85 \pm 3.62$ & $\boldsymbol{175.46} \pm 2.86$ & $566.36 \pm 3.75$ & $1810.30 \pm 0.00$ \\
    Rosenbrock & $\boldsymbol{8.92} \pm 0.33$ & $26.14 \pm 0.87$ & $76.13 \pm 0.84$ & $897.04 \pm 120.68$ \\
    \bottomrule
  \end{tabular}
  \vspace{-10pt}
\end{table}

\begin{figure}[t]
    \centering
    \begin{subfigure}[b]{0.48\textwidth}
        \centering
        \includegraphics[width=1.0\textwidth]{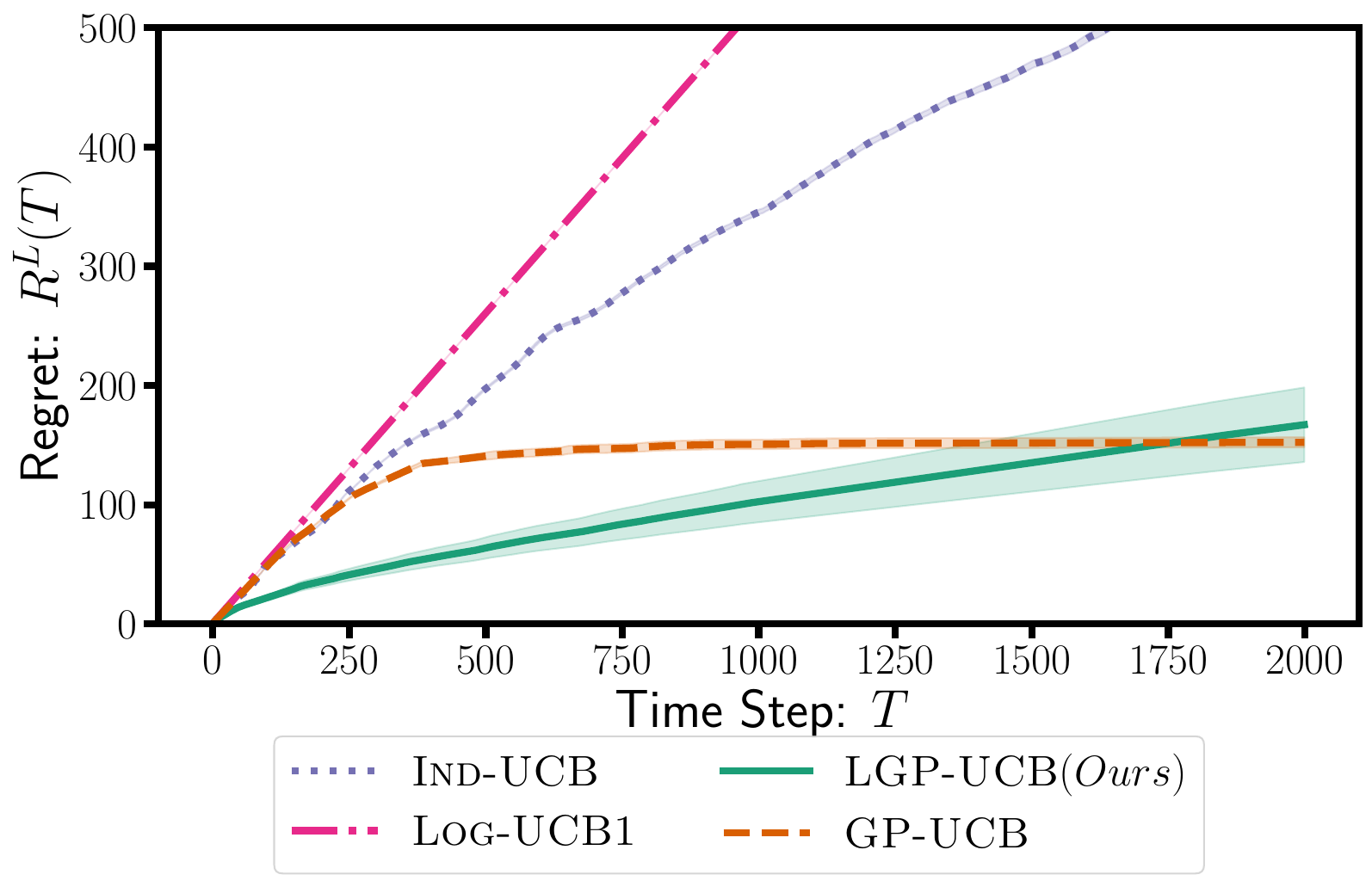}  
    \end{subfigure}
    \begin{subfigure}[b]{0.505\textwidth}
        \centering
        \includegraphics[width=1.0\textwidth]{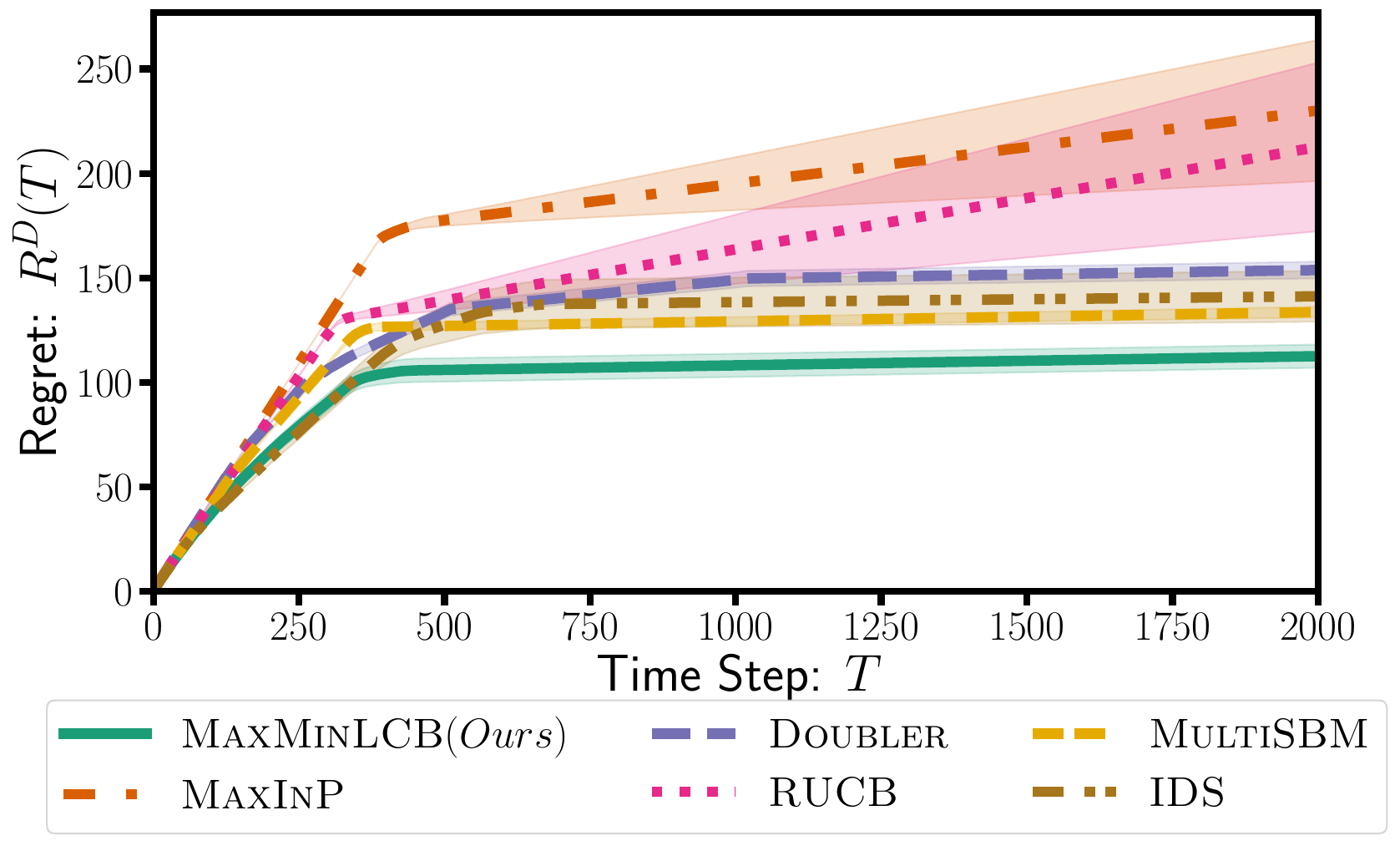}
       
    \end{subfigure}
     \vspace{-20pt}
\end{figure}
\begin{figure}[h]
    \centering
    \begin{subfigure}[b]{0.48\textwidth}
        \centering
        \includegraphics[width=1.0\textwidth]{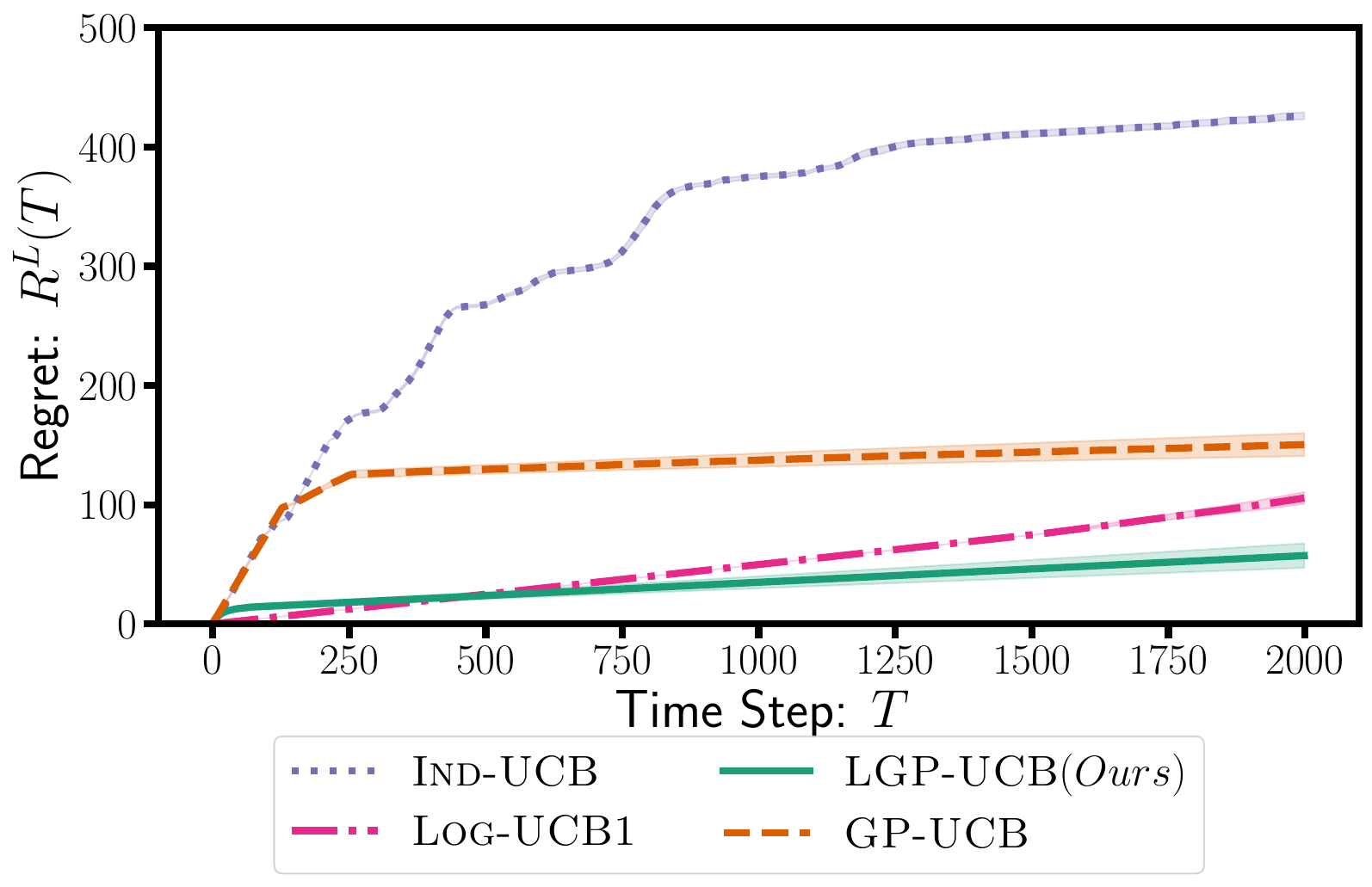}   
            
    \end{subfigure}
    \begin{subfigure}[b]{0.505\textwidth}
        \centering
        \includegraphics[width=1.0\textwidth]{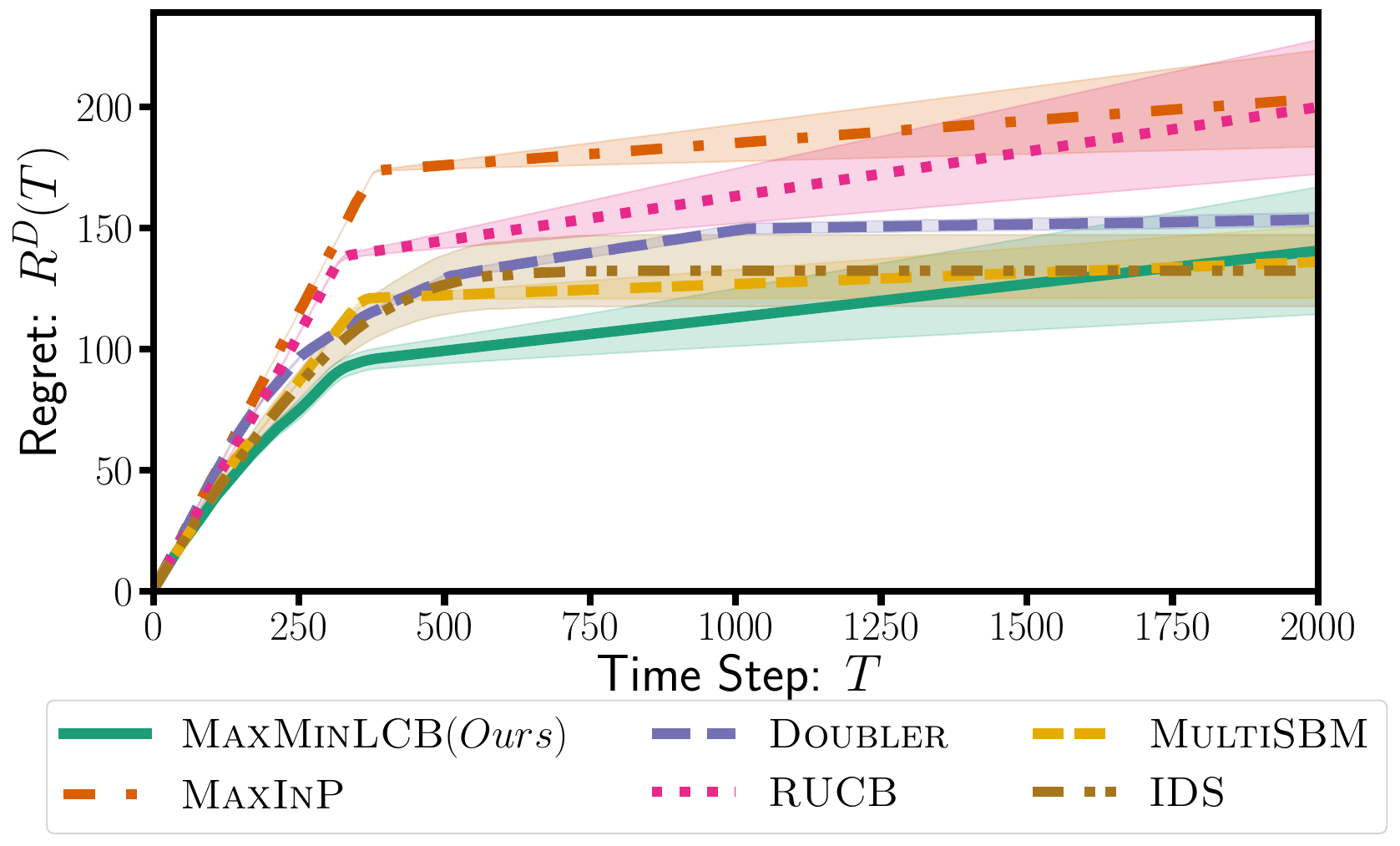}
    \end{subfigure}
     \vspace{-20pt}
\end{figure}

\begin{figure}[h]
    \centering
    \begin{subfigure}[b]{0.48\textwidth}
        \centering
        \includegraphics[width=1.0\textwidth]{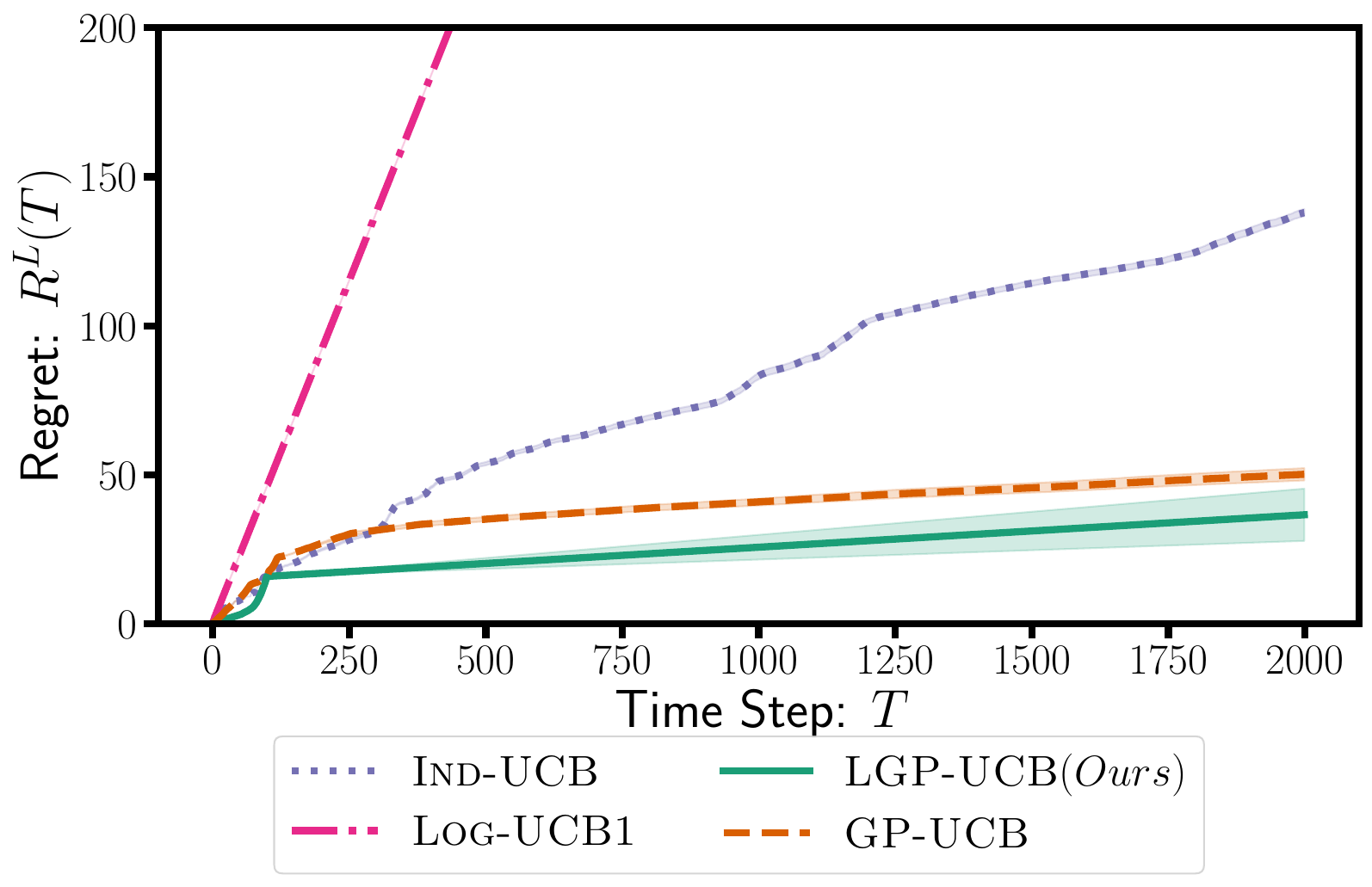}   
            
    \end{subfigure}
    \begin{subfigure}[b]{0.505\textwidth}
        \centering
        \includegraphics[width=1.0\textwidth]{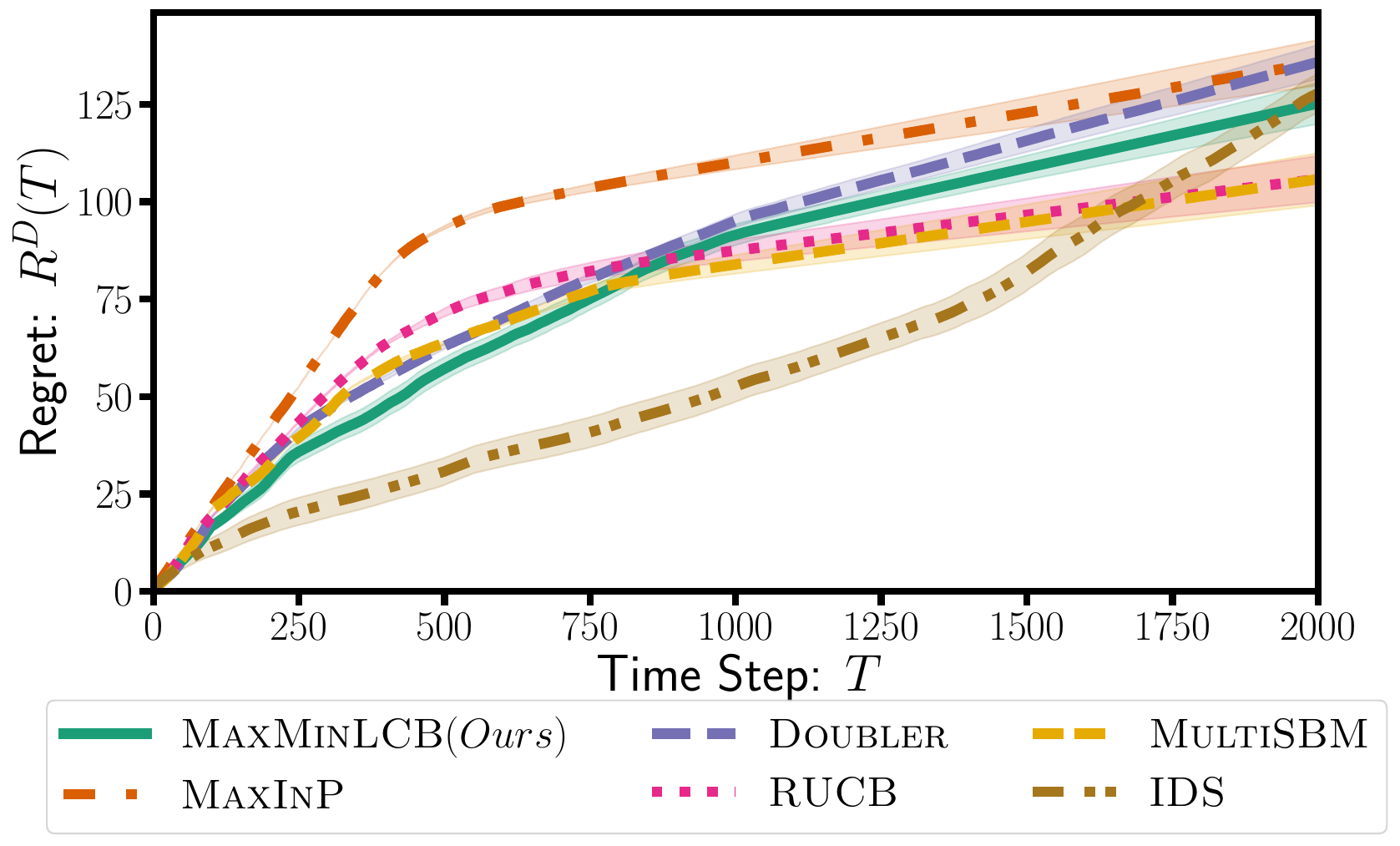}
       
    \end{subfigure}
     \vspace{-20pt}
\end{figure}

\begin{figure}[h]
    \centering
    \begin{subfigure}[b]{0.48\textwidth}
        \centering
        \includegraphics[width=1.0\textwidth]{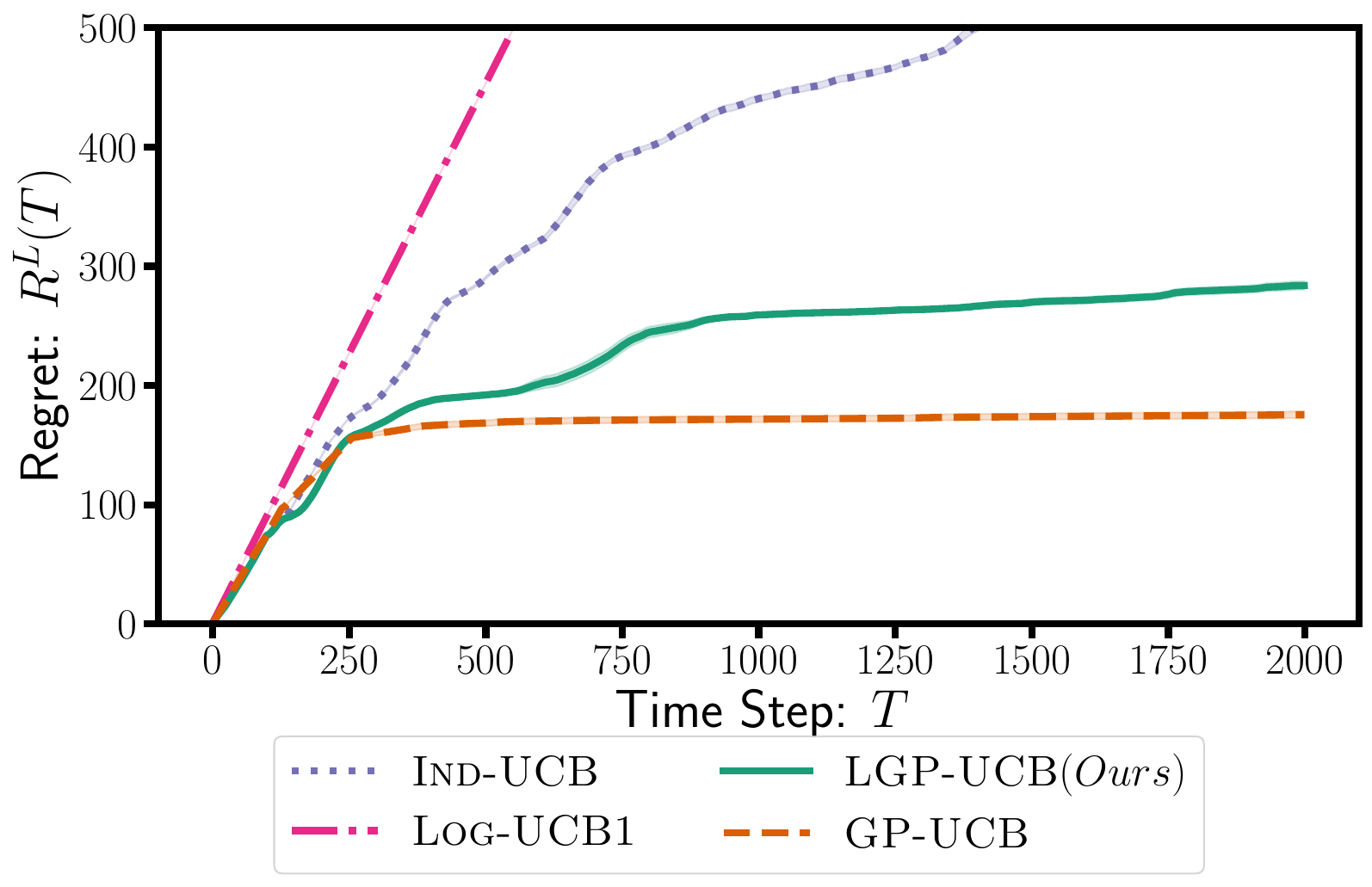}   
            
    \end{subfigure}
    \begin{subfigure}[b]{0.505\textwidth}
        \centering
        \includegraphics[width=1.0\textwidth]{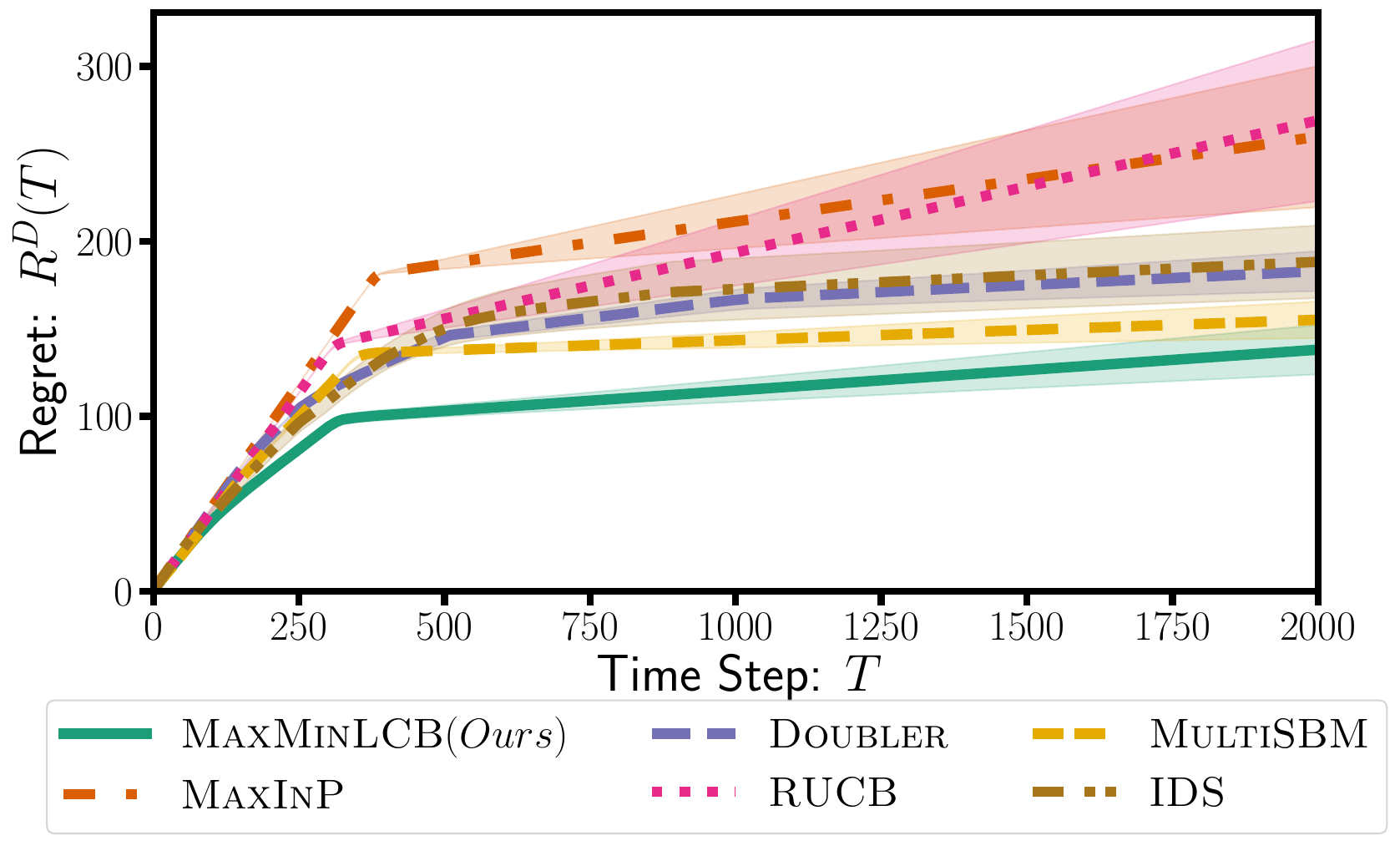}
       
    \end{subfigure}
     \vspace{-10pt}
\end{figure}

\setcounter{figure}{4}
\begin{figure}[h]
    \centering
    \begin{subfigure}[b]{0.48\textwidth}
        \centering
        \includegraphics[width=1.0\textwidth]{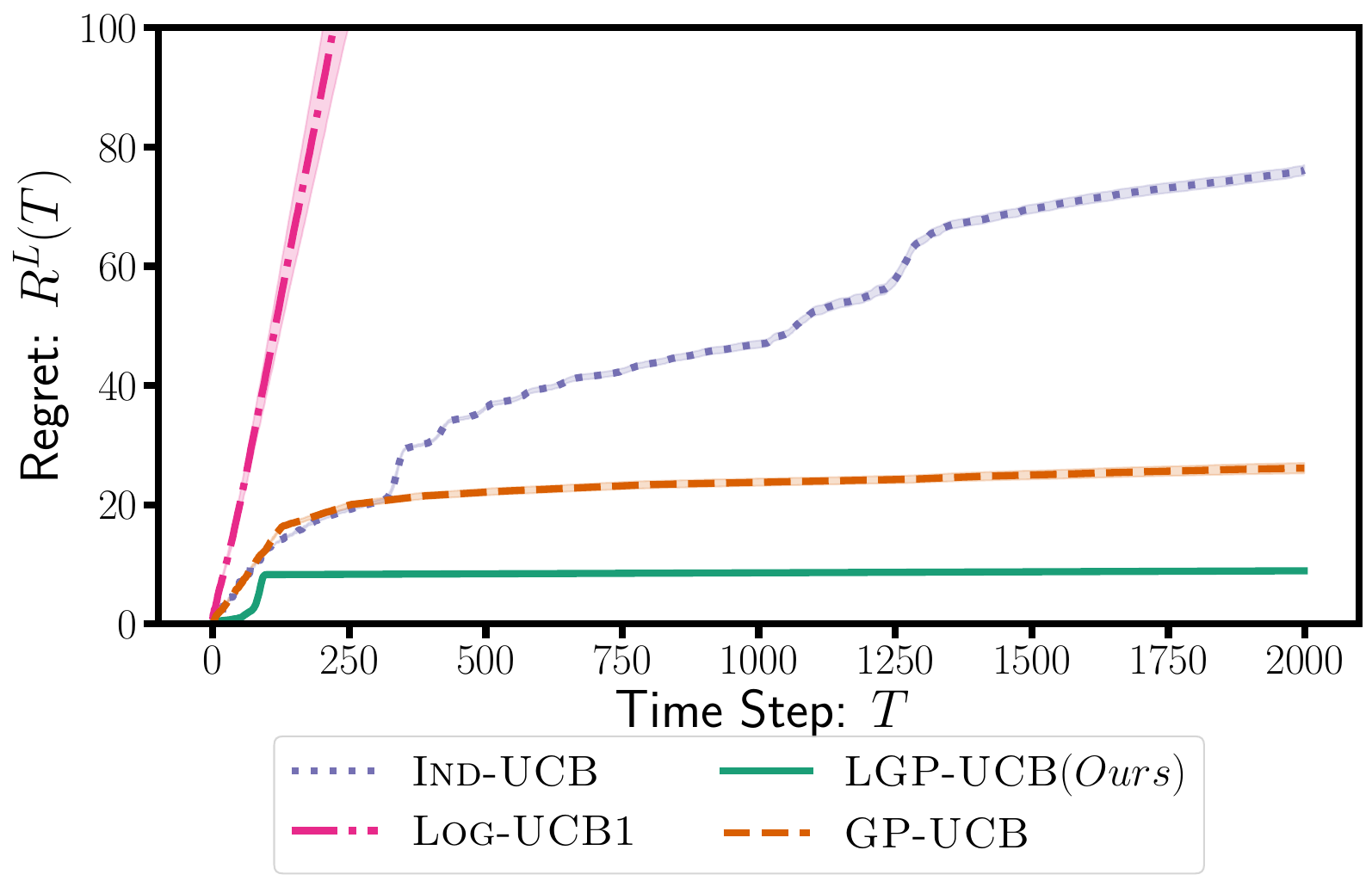}   
            
    \end{subfigure}
    \begin{subfigure}[b]{0.505\textwidth}
        \centering
        \includegraphics[width=1.0\textwidth]{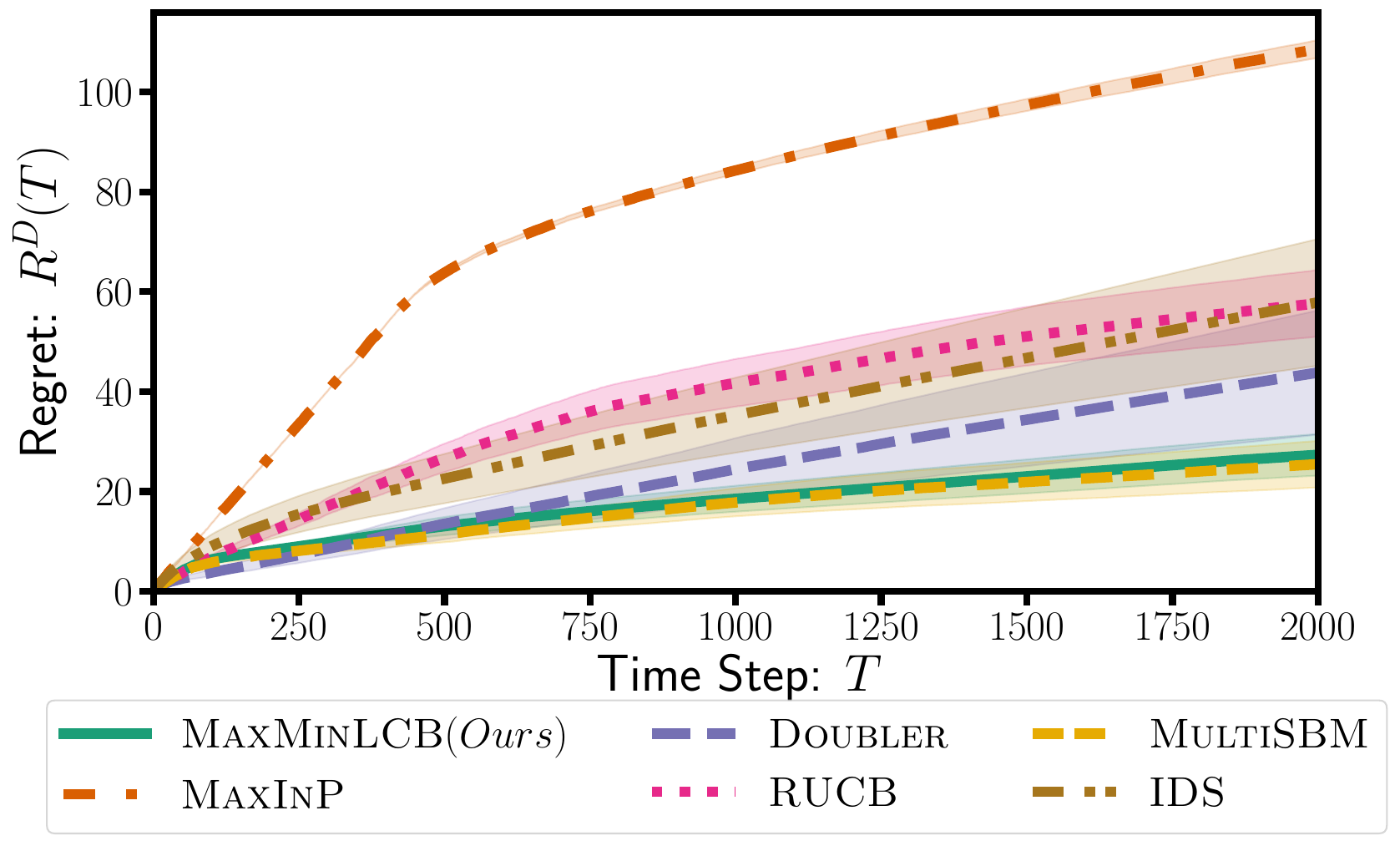}
       
    \end{subfigure}
    \caption{Top to bottom Regret for Eggholder, Hölder, Matyas Michalewicz, Rosenbrock functions, with logistic (left) and preference (right) feedback.
    \label{fig:all}}
     \vspace{-20pt}
\end{figure}